\documentclass[letterpaper, 10 pt, conference]{ieeeconf}  

\IEEEoverridecommandlockouts                              

\let\labelindent\relax
\usepackage{enumitem}

\usepackage{amsmath}
\usepackage{graphicx}
\usepackage{duckuments}
\usepackage{subcaption}
\usepackage{gensymb}
\usepackage{booktabs}
\usepackage{tabularx}
\usepackage{hyperref}
\usepackage{cleveref}
\usepackage{float} 

\title{\LARGE \bf
RobotCycle: Assessing Cycling Safety in Urban Environments
}

\author{Efimia Panagiotaki, Tyler Reinmund, Stephan Mouton, Luke Pitt, Arundathi Shaji Shanthini, Wayne Tubby, \\ Matthew Towlson,  Samuel Sze, Brian Liu,  Chris Prahacs, Daniele De Martini, Lars Kunze \\
Oxford Robotics Institute, Department of Engineering Science \\ 
University of Oxford, UK
\thanks{This work was supported by a Google DeepMind Engineering Science Scholarship, the EPSRC project RAILS (grant reference: EP/W011344/1), and the Oxford Robotics Institute research project RobotCycle. Map data copyrighted OpenStreetMap contributors and available from \url{https://www.openstreetmap.org}.   
Correspondence: \texttt{\{efimia,daniele,lars,cprahacs\}@robots.ox.ac.uk}}%
}

\usepackage[acronym]{glossaries}
\newacronym{vru}{VRU}{Vulnerable Road User}
\newacronym{imu}{IMU}{Inertial Measurement Unit}
\newacronym{av}{AV}{Autonomous Vehicle}
\newacronym{hdmap}{HD Map}{High-Definition Map}

\usepackage{crossreftools}
\makeatletter
\newcommand{\optionaldesc}[2]{%
\phantomsection
#1\protected@edef\@currentlabel{#1}\label{#2}%
}
\makeatother

\usepackage{xcolor}

\begin{document}

\maketitle
\thispagestyle{empty}
\pagestyle{empty}


\begin{abstract}
This paper introduces RobotCycle, a novel ongoing project that leverages \gls{av} research to investigate how road infrastructure influences cyclist behaviour and safety during real-world journeys.
The project's requirements were defined in collaboration with key stakeholders, including city planners, cyclists, and policymakers, informing the design of risk and safety metrics and the data collection criteria. We propose a data-driven approach relying on a novel, rich dataset of diverse traffic scenes and scenarios captured using a custom-designed wearable sensing unit. 
By analysing road-user trajectories, we identify normal path deviations indicating potential risks or hazardous interactions related to infrastructure elements in the environment. 
Our analysis correlates driving profiles and trajectory patterns with local road segments, driving conditions, and road-user interactions to predict traffic behaviours and identify critical scenarios.
Moreover, by leveraging advancements in \gls{av} research, the project generates detailed 3D \glspl{hdmap}, traffic flow patterns, and trajectory models to provide a comprehensive assessment and analysis of the behaviour of all traffic agents.
These data can then inform the design of cyclist-friendly road infrastructure, ultimately enhancing road safety and cyclability. The project provides valuable insights for enhancing cyclist protection and advancing sustainable urban mobility.
\end{abstract}


\glsresetall

\section{Introduction}
\label{sec:introduction}

\begin{figure}[t]
  \centering
  \hfill
  \begin{subfigure}{0.46\columnwidth}
    \centering
    \includegraphics[width=\columnwidth]{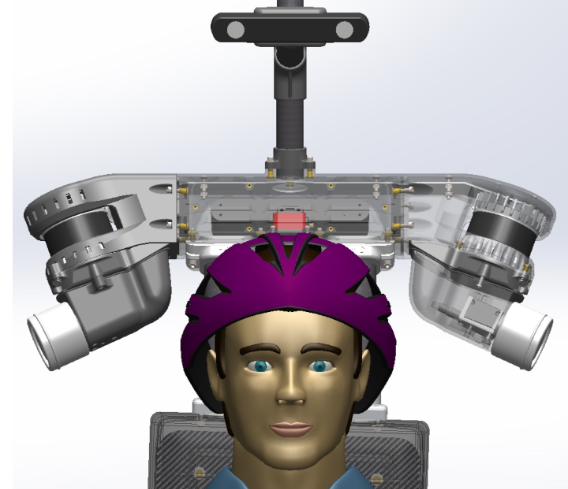}
  \end{subfigure}
  \hfill
  \begin{subfigure}{0.46\columnwidth}
    \centering
    \includegraphics[width=\columnwidth]{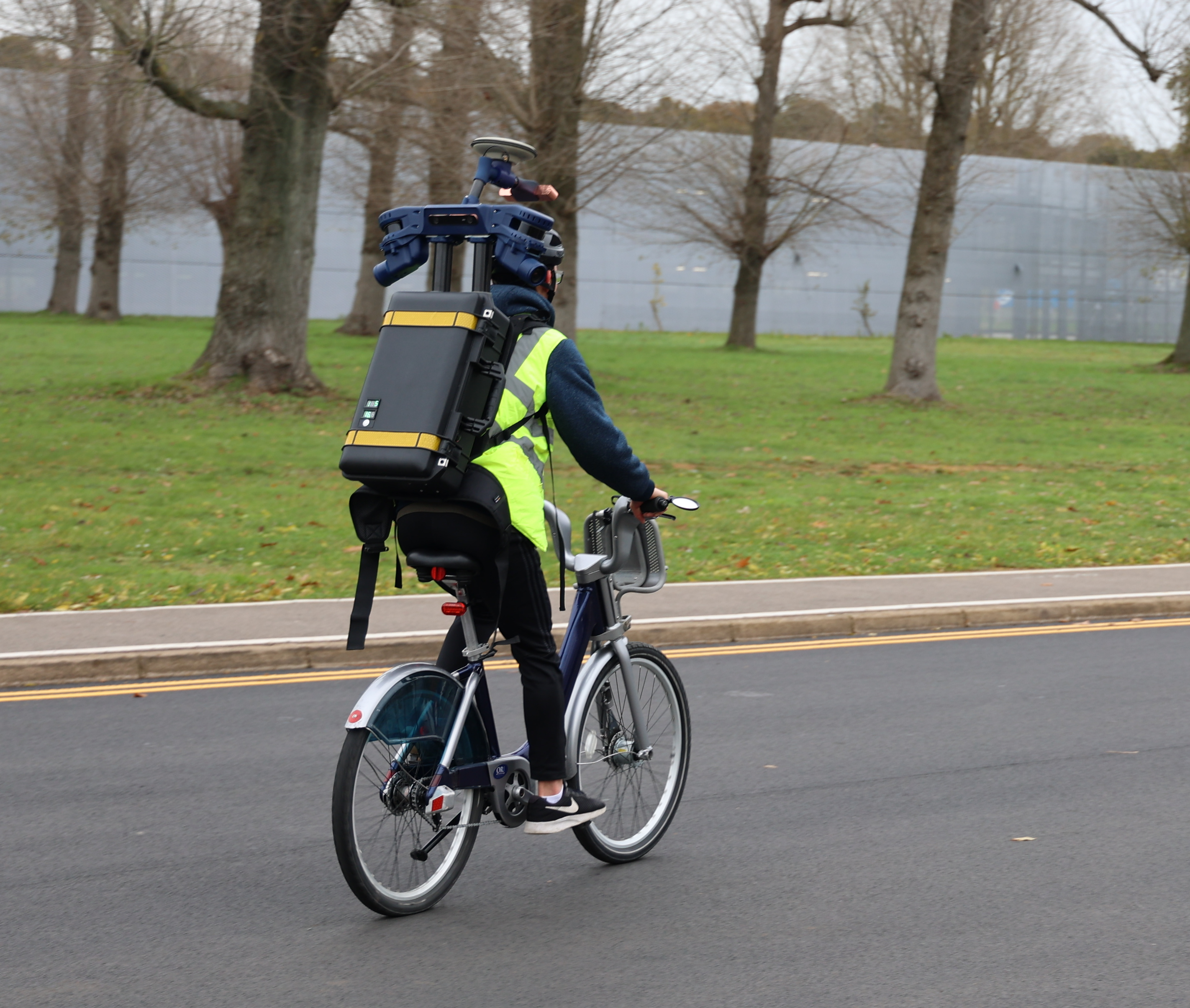}
  \end{subfigure}
  \hfill
  
  \vspace{10pt} 
  
  \hfill
  \begin{subfigure}{0.46\columnwidth}
    \centering
    \includegraphics[width=\columnwidth]{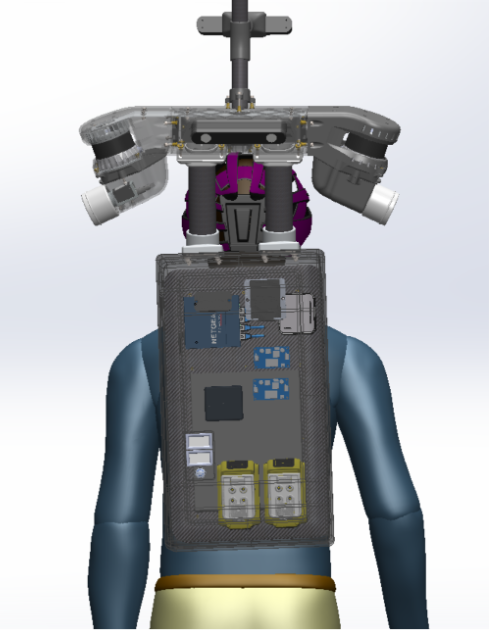}
    \caption{}
    \label{fig:sub3}
  \end{subfigure}
  \hfill
  \begin{subfigure}{0.46\columnwidth}
    \centering
    \includegraphics[width=\columnwidth]{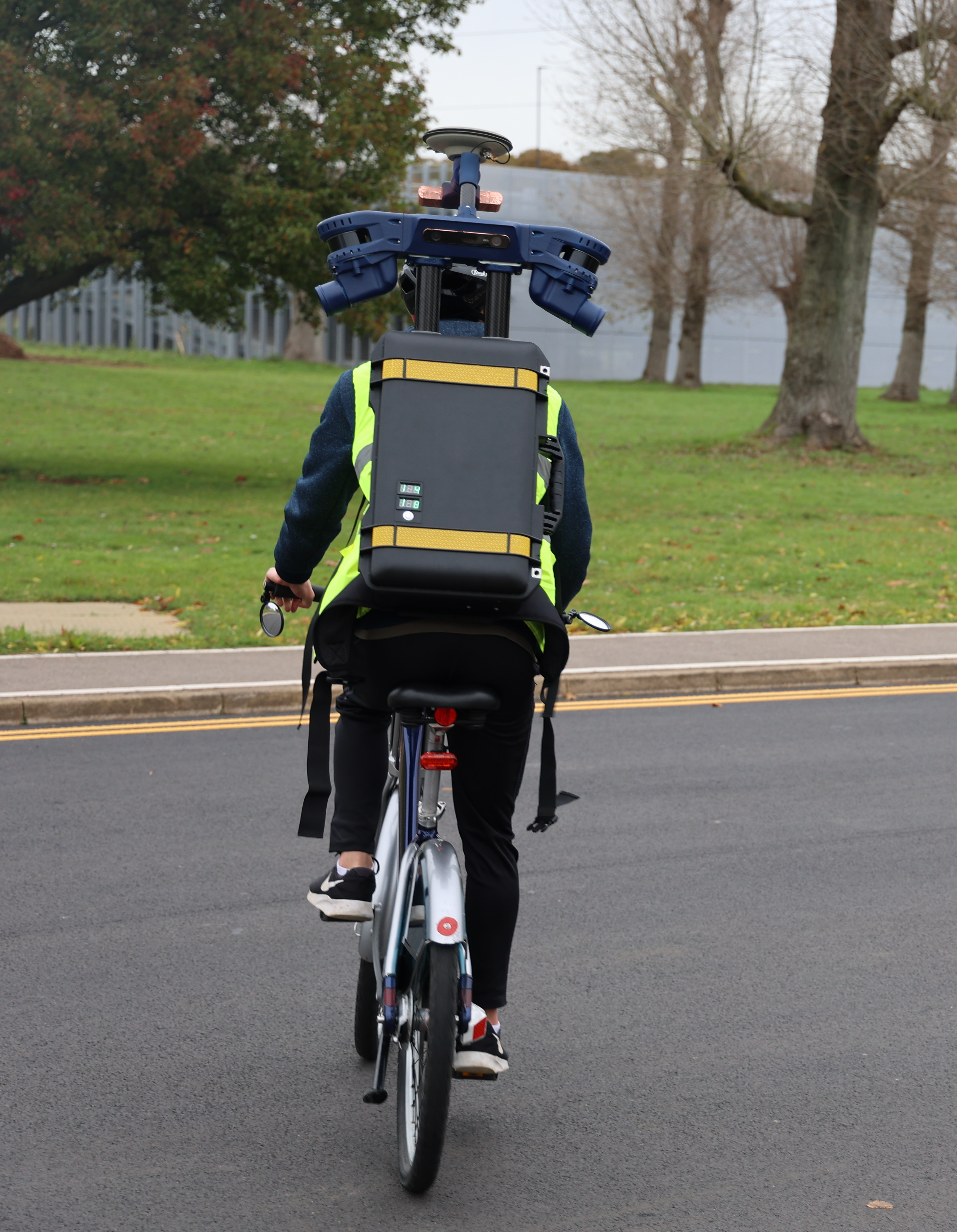}
    \caption{}
    \label{fig:sub4}
  \end{subfigure}
  \hfill
  \caption{Data-collection platform design (a) and deployment (b). The sensors and computing unit have been integrated into a backpack, allowing application flexibility without compromising the specifications and FOVs of the sensors.}
  \label{fig:main}
  \vspace{-15pt}
\end{figure}

With 15,693 cyclist casualties in UK road accidents in 2022\footnote{Reported road casualties Great Britain, annual report 2022, available at \url{http://tinyurl.com/5n8zva74}}, including fatalities and injuries, and 243 casualties in Oxfordshire alone in 2021\footnote{Oxfordshire County Council (OCC) report 2021 on Road Traffic Collisions, available at \url{http://tinyurl.com/43ff89nc}}, cyclists remain among the most \gls{vru}. Cyclists are continually exposed to risks due to the inherent design of roads, unpredictable behaviour of other users, and inadequate safety measures.

Our ongoing RobotCycle project aims to address these challenges by identifying how specific road features, such as road layout and traffic filters, along with the behaviour of other road users, can impact cyclist safety in urban environments. 
Incident reports and accident statistics lack crucial contextual information surrounding critical events since they mainly rely on post-event reporting and stationary sensing. These methods provide limited contextual information from a restricted set of sensors configured in fixed locations and viewpoints.

RobotCycle complements these approaches by gathering detailed information about traffic, road infrastructure, and behavioural patterns from the perspective of the \glspl{vru}, analysing all the factors contributing to cyclist vulnerability.
We equip cyclists with a mobile sensing unit incorporating state-of-the-art technologies commonly found in \glspl{av} and further combine these rich ego-centric data with \glspl{hdmap}, trajectory tracking, and cyclist feedback to microscopically investigate the causes of safety-critical incidents and determine how road infrastructure and users' behaviour impact cyclists' safety during complex social interactions on urban roads.
\Cref{fig:main} shows the final design of the sensorised backpack.
We aim to identify previously unrecognized safety issues as well as risky and complex behavioural patterns. Through this, we seek to inform the development of more effective interventions and deepen our understanding of social interactions in real-world road contexts, effectively enhancing cyclist safety.

\section{Structure and Objectives}

The primary scope of the RobotCycle project is to investigate how cycling infrastructure influences cyclist behaviour and safety \textit{during real-world journeys} to inform the design of cyclist-friendly road layouts, improving road safety and cyclability.
%
Through our engagement with the key stakeholders, such as policymakers and cyclist associations, we formed the following three objectives:
\begin{description}[itemindent=0pt, leftmargin=7pt]
    \item[{\crtcrossreflabel{\textcolor{blue}{O1}}[objective.backpack]}.] Design a novel, wearable sensing unit to collect rich multi-sensor data from the perspective of the cyclists;
    \item[{\crtcrossreflabel{\textcolor{blue}{O2}}[objective.dataset]}.] Develop a novel, annotated dataset for analysing the factors contributing to road safety, focusing on road infrastructure and traffic interactions;
    \item[{\crtcrossreflabel{\textcolor{blue}{O3}}[objective.benchmark]}.] Propose a benchmark for assessing road safety for cyclists in urban environments through safety and cyclability indexes, identifying areas of high risk.
\end{description}
In the remainder of the paper, we discuss each objective further and describe our approach to achieving them.



\section{Background and Related Work}
\label{sec:related_work}

\textbf{Cyclist Journey Analysis. }
Recent studies investigate cycling safety and road users' interactions by modelling driving, riding, and walking behaviours. These projects leverage various data sources to capture the environment, including sensors directly mounted on bicycles or cyclists and personalised IoT cycling gear \cite{Shoman2023, Kaparias2021, Vieira2016, Ibrahim2021a}. 
For instance, \cite{Shoman2023} conducts a thorough analysis, considering social and kinematic factors to evaluate cyclists' safety and comfort in challenging environmental conditions. Capturing also social characteristics, \cite{Kaparias2021} assesses highway design and its impact on cycling while \cite{Kumar2023} relies on IoT cycling gear for personalised cyclist training and safety assessment. Additionally, \cite{Vieira2016} captures stress and health data from cyclists and correlates them with image data, focusing on specific events.
\cite{Ibrahim2021a} proposes a method for detecting near misses from video captured from various positions, either on a bicycle or on the cyclist's helmet. RobotCycle extends upon these works relying on wearable sensing, and capturing traffic interactions while mapping and modelling road infrastructure.

\textbf{Backpack design. }
Due to their operational flexibility, backpacks have been previously employed for mapping and data collection projects, relying either solely on lidar scanners for capturing the surroundings \cite{Wen2016, Gong2021, Chen2021, Nuchter2015, Laguela2018} or also on cameras \cite{Corso2013, Holmgren2017}. Aiming to collect a rich multi-modal dataset, we designed and built a custom versatile backpack equipped with lidars, cameras, and inertial sensors achieving a complete $360\degree$  Field-Of-View (FOV) and close-to-far range detection around the cyclist, capturing geometry and appearance.
This setup allows us to map the environment, capture close-range interactions, and analyse the trajectory of each participating actor in a traffic scene.

\textbf{Behaviour, Risk, and Safety Modelling. }
Various studies focus on modelling traffic interactions and risk, while others propose alternative infrastructure to design safer road networks.
For instance, \cite{Rasch2022} employs behaviour modelling to predict collisions and recommend avoidance manoeuvres. 
Meanwhile, \cite{Daraei2021} develops a safety model based on accident reports and mapping data to unveil behavioural patterns.
In a different approach, \cite{Castells-Graells2020} leverages risk and discomfort estimation models to generate personalised route recommendations by correlating accident and traffic data with street network topology.
Additionally, \cite{Meuleners2023} utilises simulation data to propose improvements in road infrastructure, while \cite{Ibrahim2021b} leverages adversarial networks to redesign urban cities. 
In contrast, RobotCycle adopts a data-driven approach. We collect an extensive dataset capturing diverse road layouts, infrastructure characteristics, and weather and lighting conditions. Leveraging methodologies from \gls{av} research, we generate traffic, behaviour, and risk models. These models, in correlation with detailed \glspl{hdmap}, enhance our understanding of road users' interactions facilitating the identification of incident hotspots.

\section{Wearable Sensor Unit Design}
\label{sec:backpack}

\begin{table}[]
\centering
\renewcommand{\arraystretch}{1.4}
\begin{tabular}{|l|c|ll|}
\hline
\textbf{Sensor} & \textbf{Qty} & \textbf{Model}     & \textbf{Description}                   \\ \hline
Lidar           & 2            & PandarXT-32         & 360x31° (HxV) FOV  \\
Stereo          & 2            & ZED2i              & Stereo camera, intergrated IMU \\
Camera          & 2            & Basler acA1920     & Colour monocular, fisheye lenses          \\
INS             & 1            & Ellipse-N          & RTK GNSS                           \\ \hline
\end{tabular}
\caption{Overview of the sensors used in the final version of the data collection platform. In post-processing, we utilise the fitted \glspl{imu} to mitigate noise caused by the cyclists' body movements and road vibrations.}
\label{tab:Sensors}
\vspace{-10pt}
\end{table}

Our objective \ref{objective.backpack} is to develop a comprehensive data-collection sensing unit incorporating state-of-the-art range and image sensors with inertial navigation systems. We aim to collect a robust dataset capturing short- and long-range infrastructure details and dynamic driving interactions in each traffic scene.
This holistic approach ensures that the resulting dataset can be used to analyse and understand the intricacies of various driving scenarios.

We can summarise the requirements for the proposed sensorised platform with the following:
\begin{description}[itemindent=0pt, leftmargin=7pt]
    \item[{\crtcrossreflabel{\textcolor{blue}{R1.1}}[req.backpack.weight]}.] Develop a portable, robust, and weather-invariant lightweight sensing unit to be used by different \glspl{vru};
    \item[{\crtcrossreflabel{\textcolor{blue}{R1.2}}[req.backpack.sensors]}.] Achieve a combined multi-modal $360\degree$ FOV short-to-long range sensing coverage, also capturing inertial and positional measurements;
    \item[{\crtcrossreflabel{\textcolor{blue}{R1.3}}[req.backpack.battery]}.] Record four hours of continuous cycling data per day for two weeks, capturing varying times of day, traffic volumes, and weather conditions;
    \item[{\crtcrossreflabel{\textcolor{blue}{R1.4}}[req.backpack.stealth]}.] Aim for inconspicuous design.
\end{description}
Importantly, project requirements are derived from and interlinked among the objectives.
Indeed, \ref{req.backpack.sensors} is related to \ref{objective.dataset}'s, \ref{req.data.diversity}, and \ref{objective.benchmark}'s analysis, and \ref{req.backpack.weight} and \ref{req.backpack.battery} to \ref{objective.dataset}'s and \ref{req.data.battery}.

\begin{figure*}[h]
    \centering
    \begin{subfigure}{0.25\textwidth}
        \includegraphics[width=\textwidth]{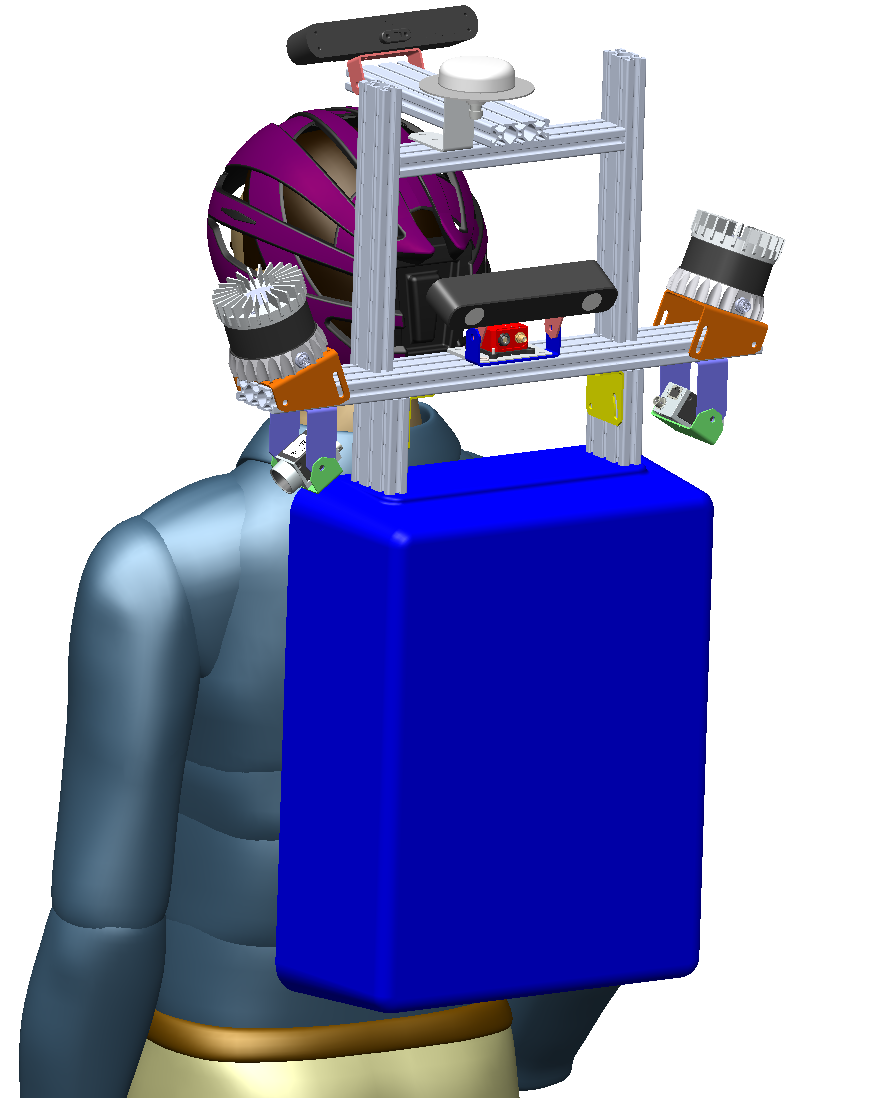}
    \end{subfigure}
    \begin{subfigure}{0.25\textwidth}
        \includegraphics[width=\textwidth]{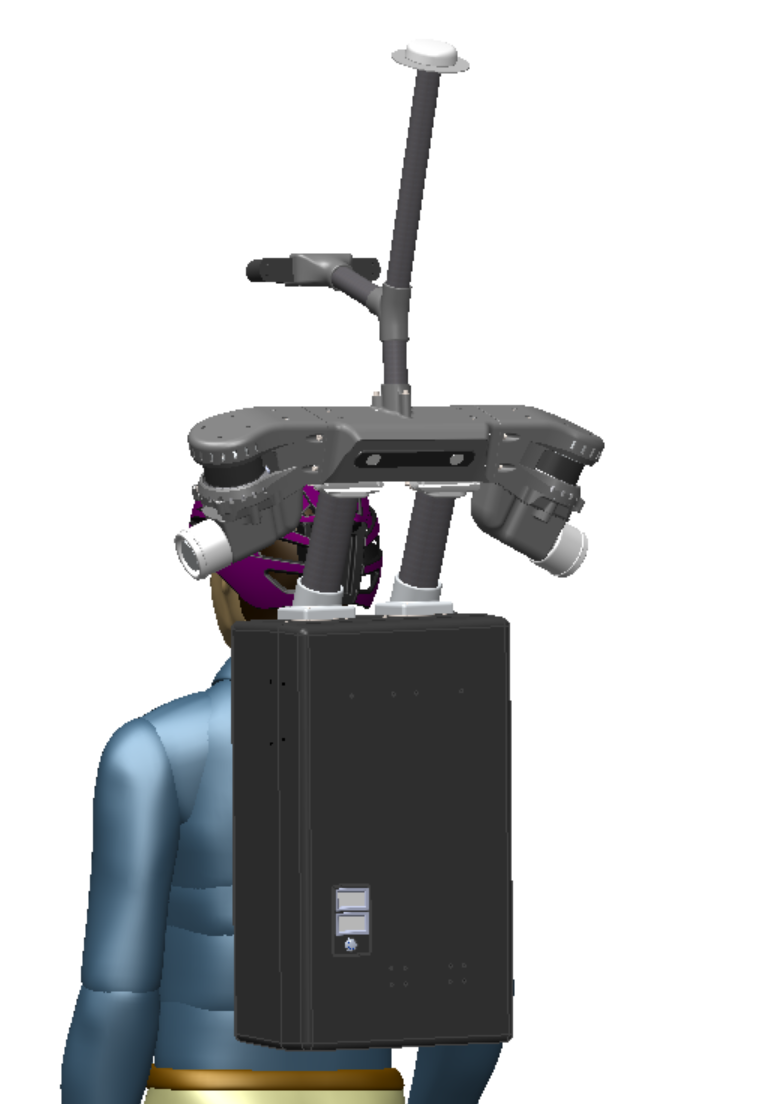}
    \end{subfigure}
    \begin{subfigure}{0.25\textwidth}
        \includegraphics[width=\textwidth]{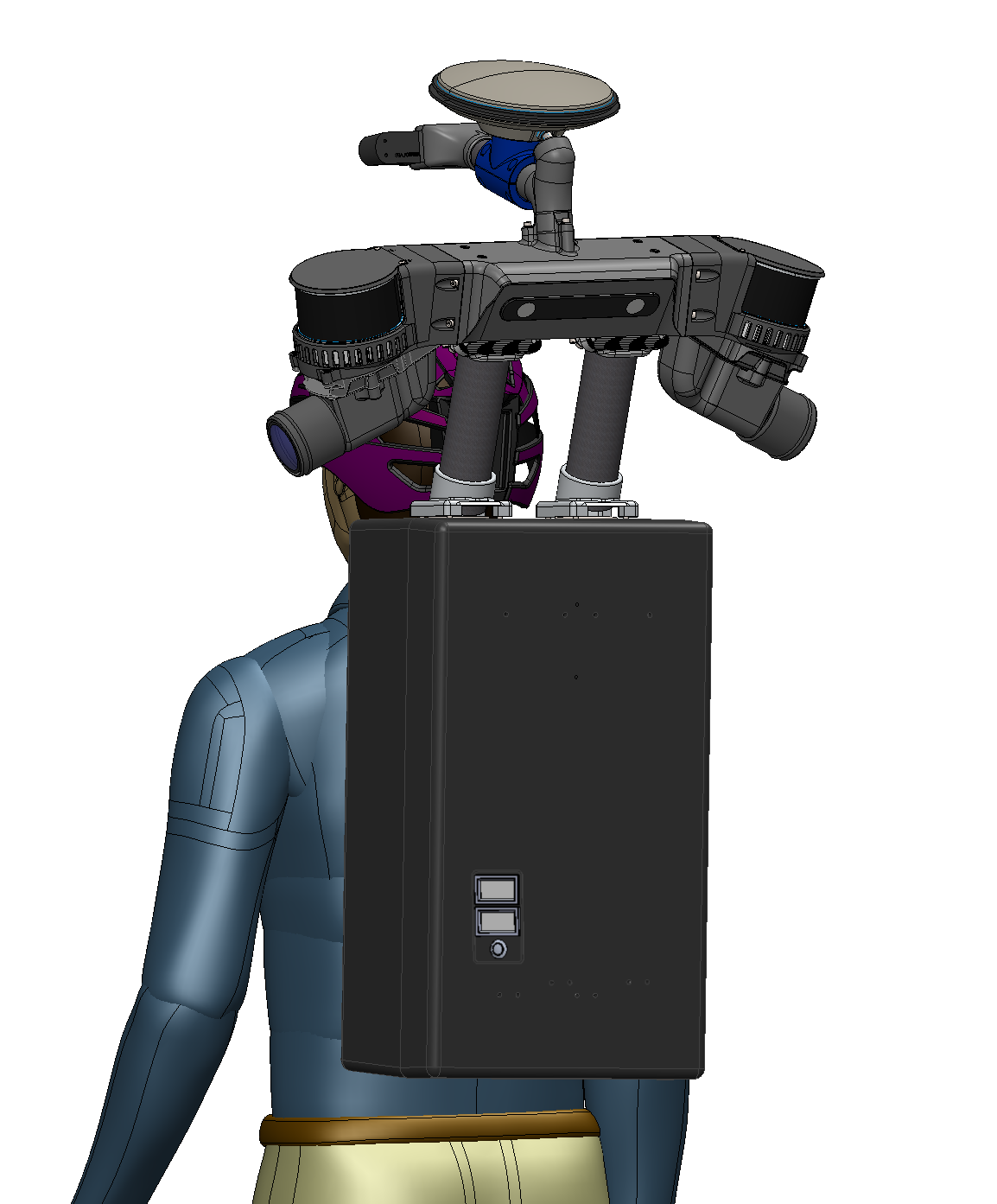}
    \end{subfigure}
    \vspace{10pt}

    \begin{subfigure}{0.25\textwidth}
        \includegraphics[width=\textwidth]{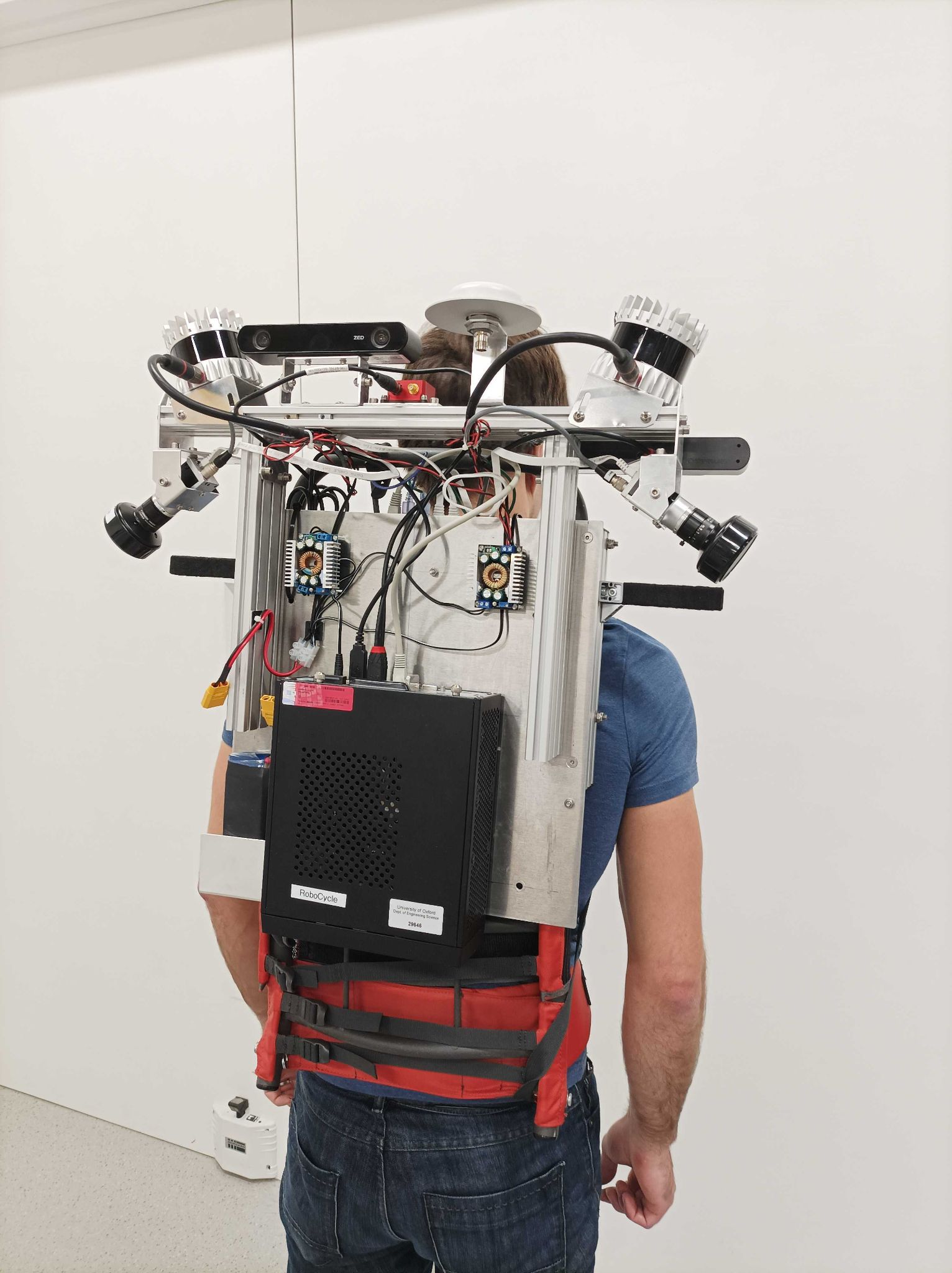}
        \caption{}
    \end{subfigure}
    \begin{subfigure}{0.25\textwidth}
        \includegraphics[width=\textwidth]{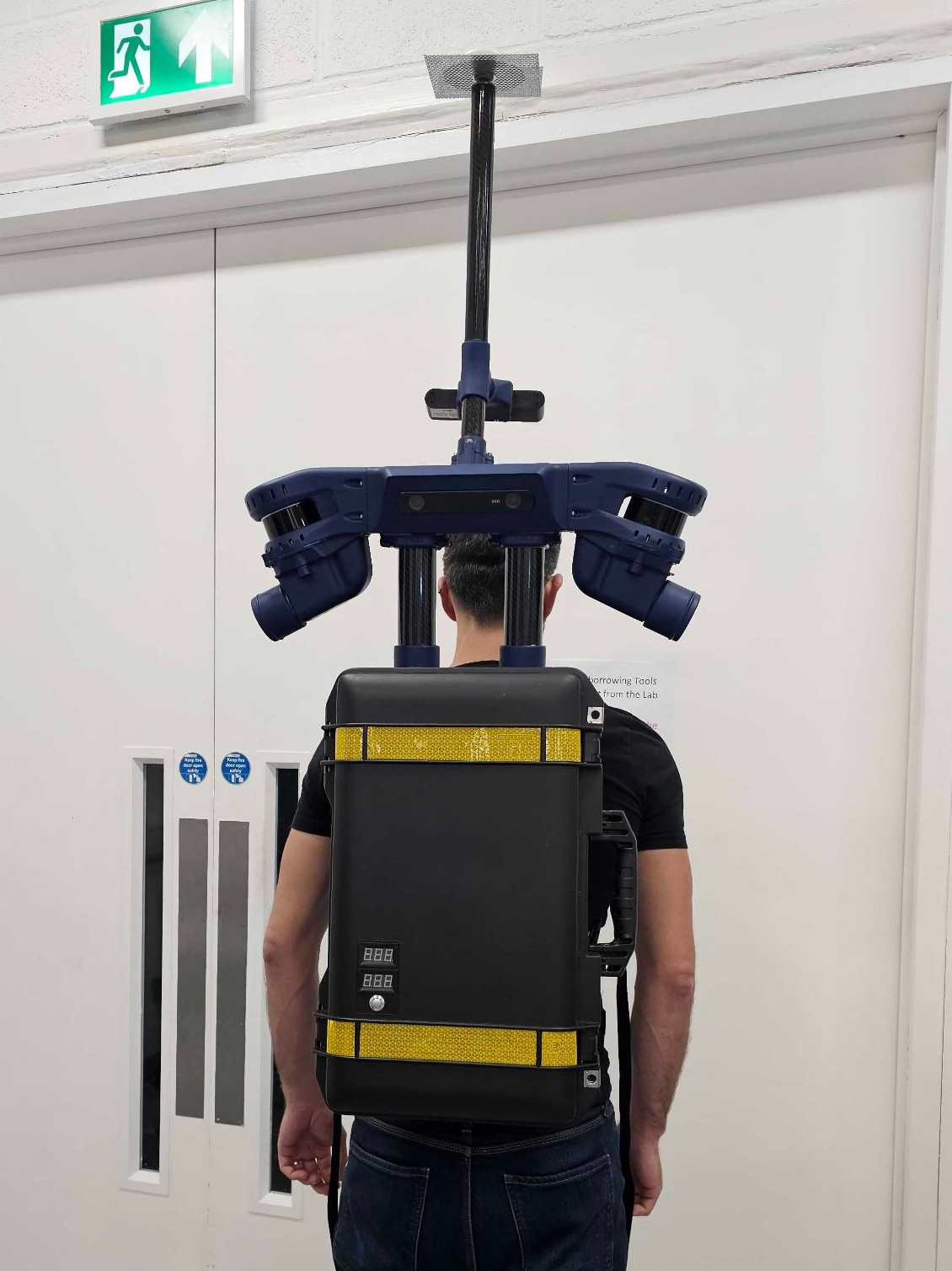}
        \caption{}
    \end{subfigure}
    \begin{subfigure}{0.25\textwidth}
        \includegraphics[width=\textwidth]{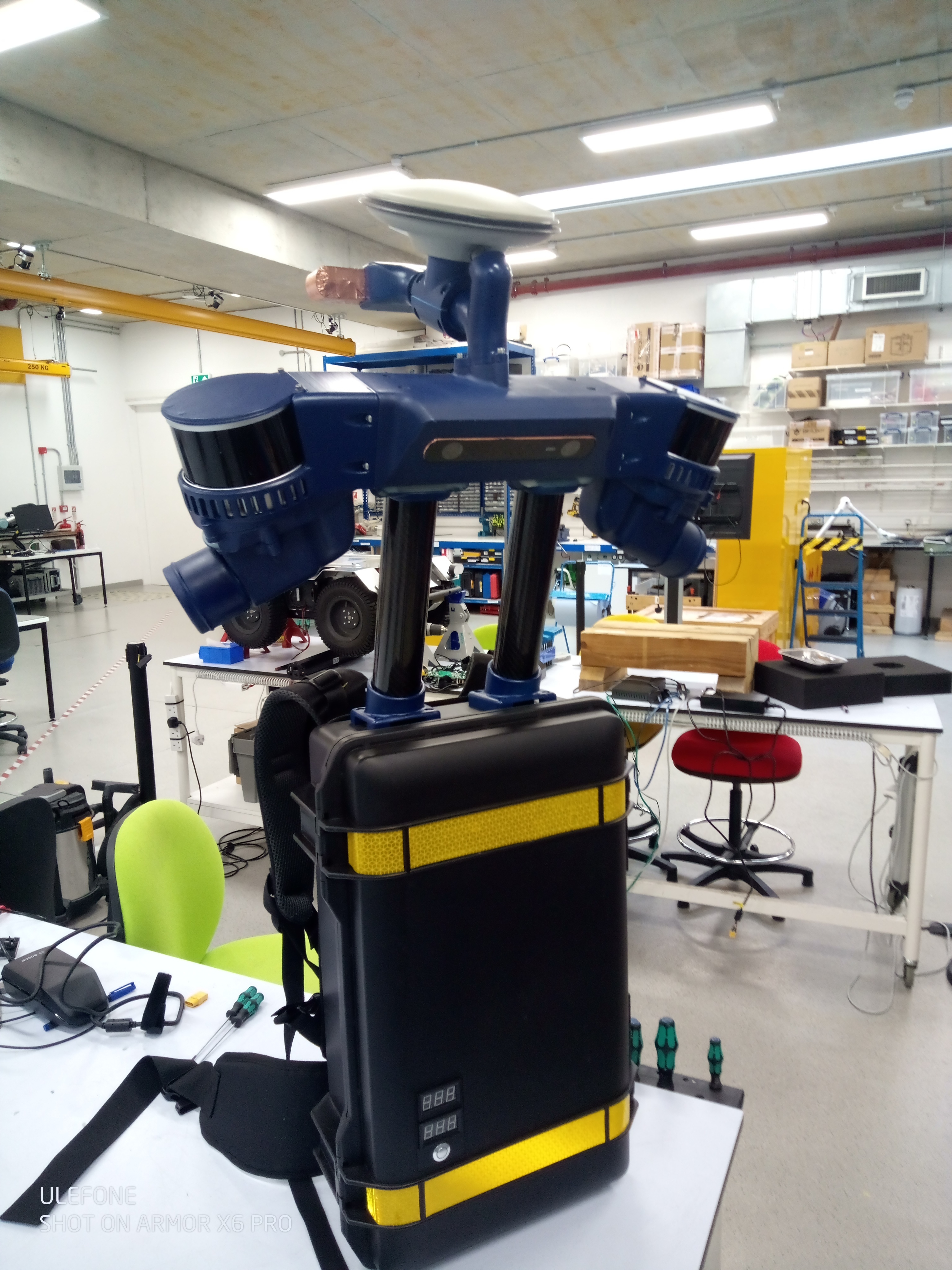}
        \caption{}
    \end{subfigure}
    
    \caption{Various design iterations of the RobotCycle backpack.
    From the left, the first iteration (a) allowed for sensors' displacement to optimise their position and orientation.
    The second iteration (b) consolidated the sensors' positioning, computing requirements, and networking, allowing extensive field testing.
    The final iteration (c), finalised the sensing configuration and design, maximising sensors' FOV and optimising overall weight distribution.}
    \label{fig:iterations}
    \vspace{-10pt}
\end{figure*}

To satisfy the design requirements, we opted for a sensorised backpack.
This wearable solution enhances deployment flexibility as it does not depend on external or fixed infrastructure, such as the bicycle frame, and can be easily worn by different \glspl{vru} at the cost of increased load on the user, trading off \ref{req.backpack.weight} and \ref{req.data.battery}.
In addition, this solution does not limit our sensor selection and mounting options while simplifying the overall wiring setup.

The sensors equipped can be seen in \Cref{tab:Sensors}, sample data are visualised in \Cref{fig:sensors}, and the final design is depicted in \Cref{fig:main}; it should be noted that we prioritised \ref{req.backpack.sensors} over \ref{req.backpack.stealth} to satisfy the data analysis procedures.
The backpack also fits an onboard computer, a hardware synchronisation system, and various storage devices, designed to meet the 4-hour runtime requirement (\ref{req.backpack.battery}) while allowing quick and easy battery replacement in the field.


\subsection{Iterative design}

\begin{figure}
    \centering
    \begin{subfigure}{0.75\columnwidth}
        \includegraphics[width=\textwidth]{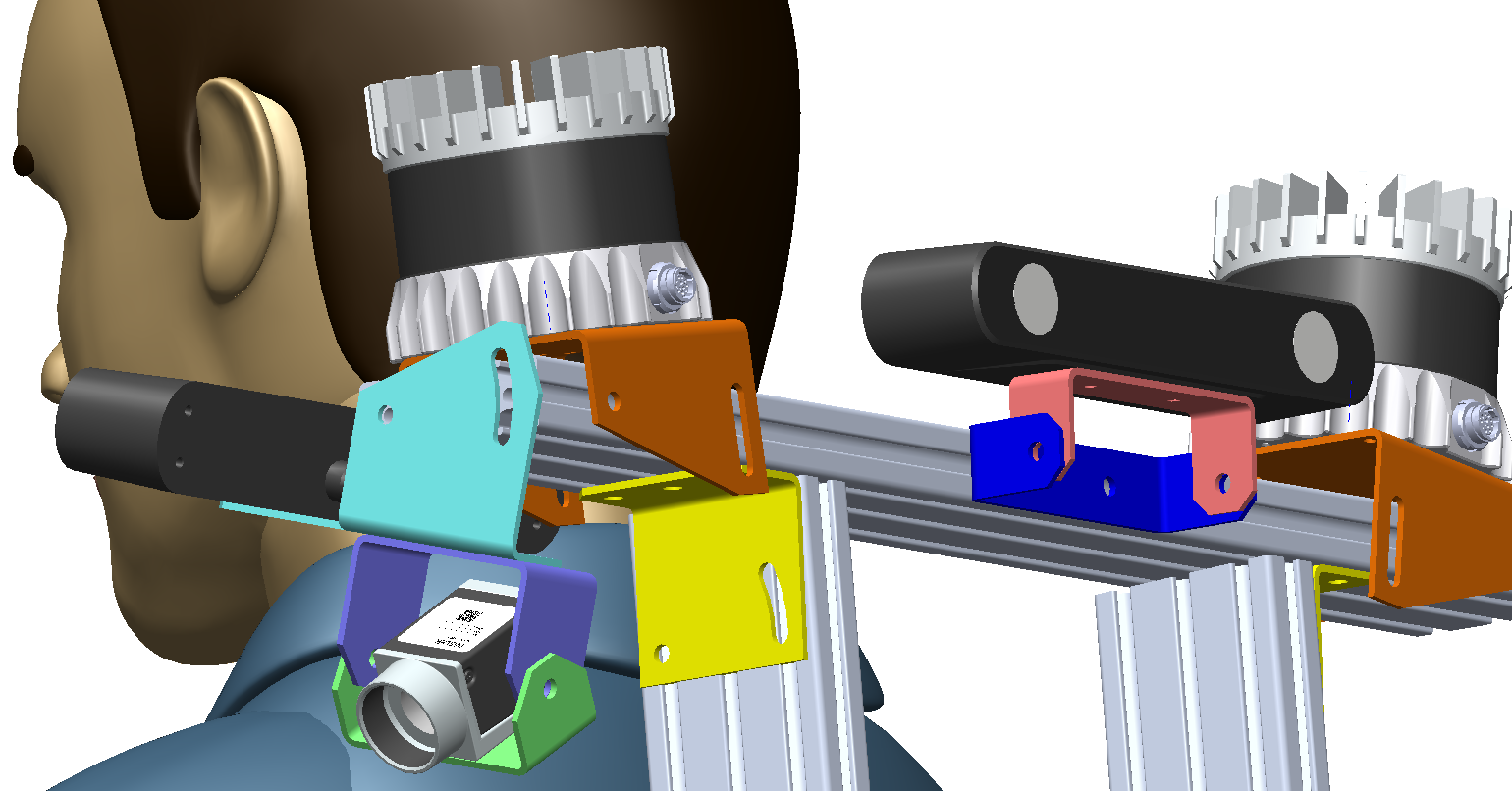}
    \end{subfigure}
    \vspace{4pt}
    
    \begin{subfigure}{0.75\columnwidth}
        \includegraphics[width=\textwidth]{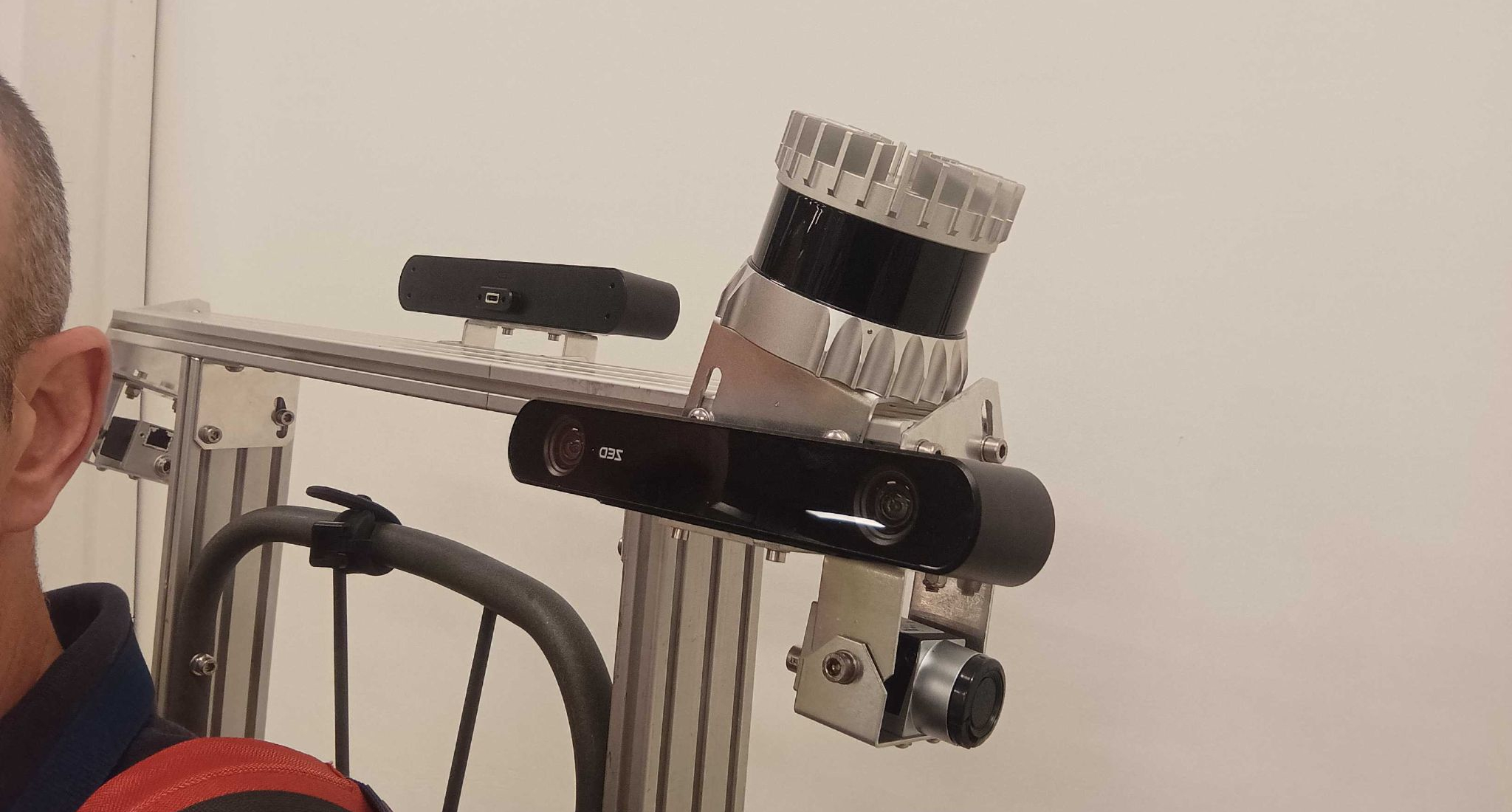}
    \end{subfigure}
    \caption{Detail of the articulated systems that enabled extensive experimentation with various sensors' positioning to optimise the sensing setup.}
    \label{fig:joints}
    \vspace{-10pt}
\end{figure}

We optimised the mounting configurations and sensors specifications to achieve an unobstructed $360\degree$ FOV coverage around the ego-cyclist after several design iterations, as seen in \Cref{fig:iterations}.
The initial design, in \Cref{fig:iterations}a, relied on custom articulated joints and mounting points, allowing us to experiment with different sensors' positioning and orientations, as well as different sensor types, specifications, and FOVs.  
\Cref{fig:joints} shows one of our custom joint mechanisms enabling six degrees of freedom movement, facilitating experimentation with various sensors' positioning. With this flexible and adjustable setup, we optimised the design, taking into account various mounting points. For instance, in \Cref{fig:joints}, the front camera is mounted on the side of the rider, illustrating one of the possible configurations we explored.

\begin{figure*}[ht]
    \centering
    \begin{subfigure}{0.3\textwidth}
        \includegraphics[width=\textwidth]{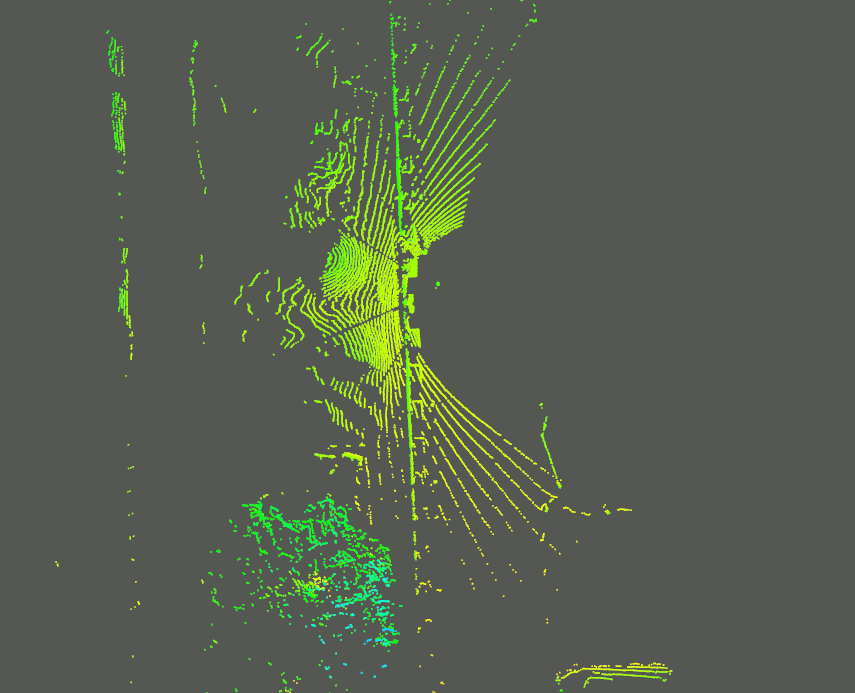}
        \caption{}
    \end{subfigure}
    \begin{subfigure}{0.3\textwidth}
        \includegraphics[width=\textwidth]{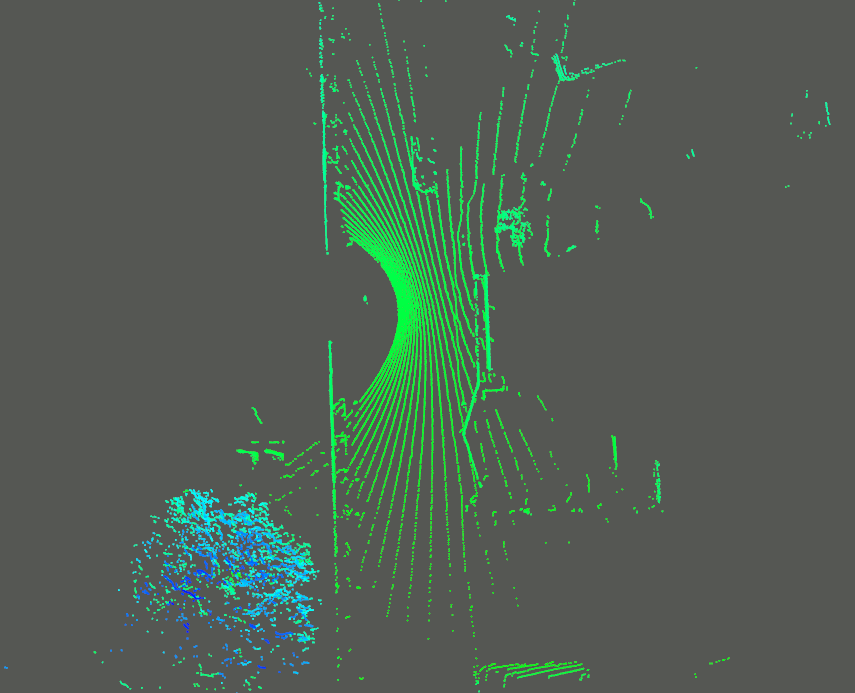}
        \caption{}
    \end{subfigure}
    \begin{subfigure}{0.3\textwidth}
        \includegraphics[width=\textwidth]{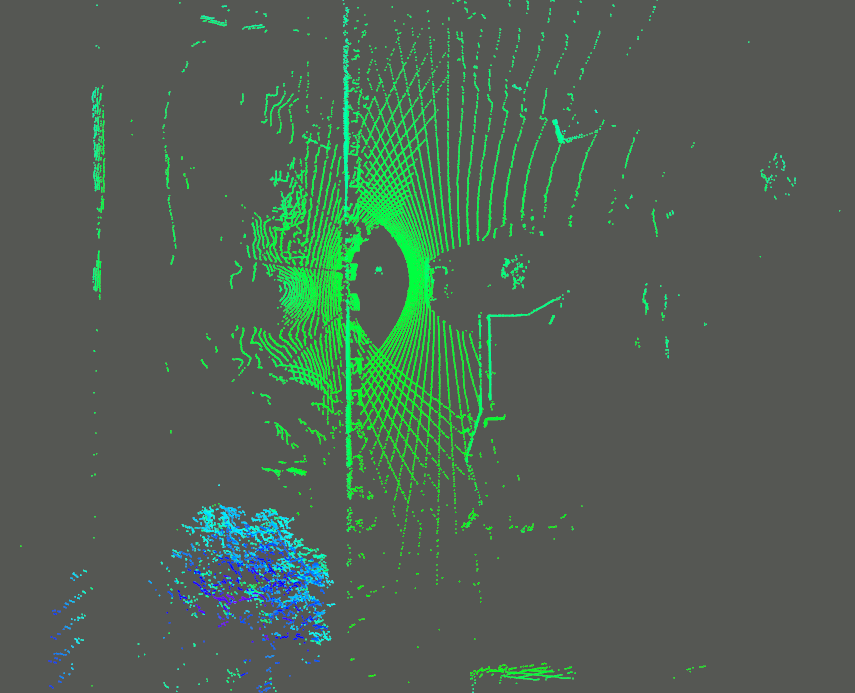}
        \caption{}
    \end{subfigure}

    \begin{subfigure}{0.6\textwidth}
        \includegraphics[width=\textwidth]{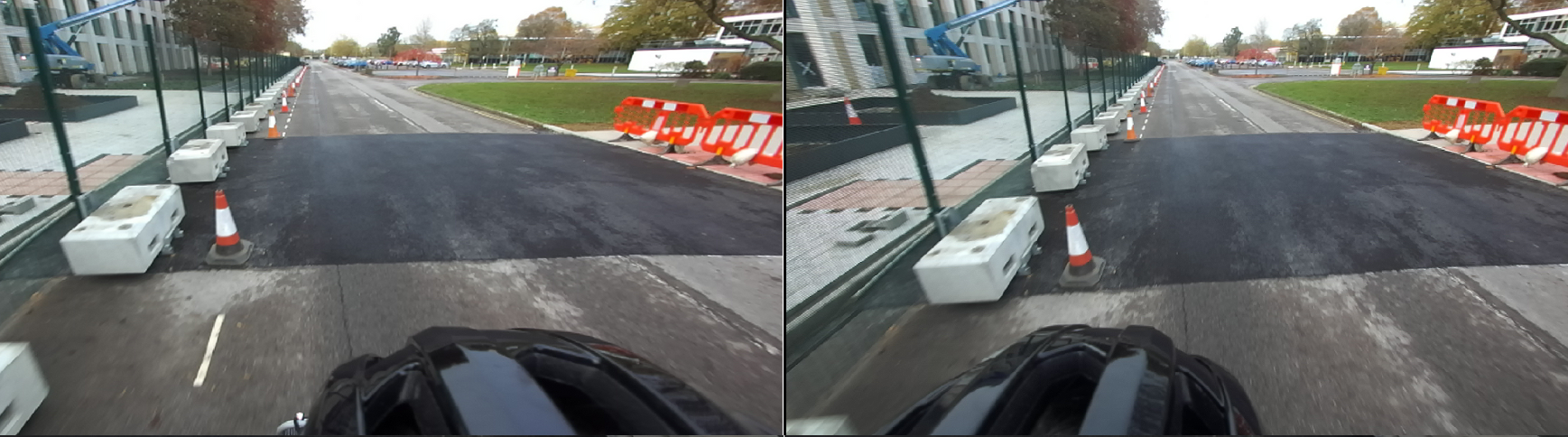}
        \caption{}
    \end{subfigure}
    \begin{subfigure}{0.3\textwidth}
        \includegraphics[width=\textwidth]{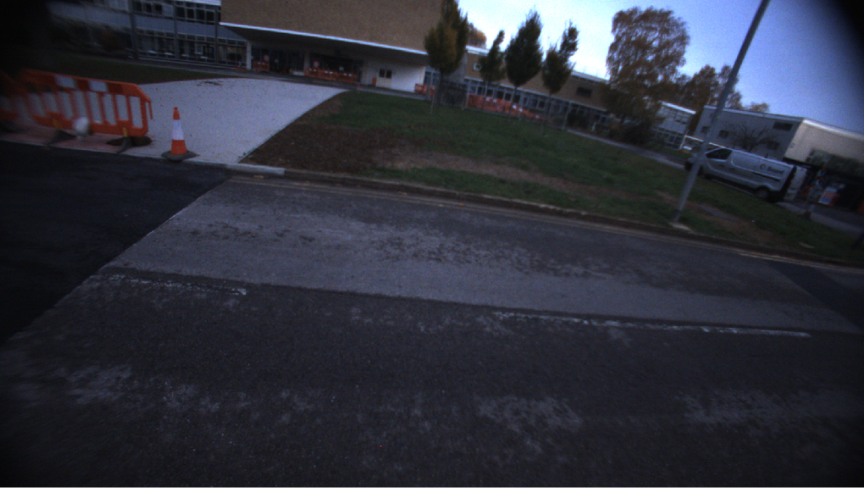}
        \caption{}
    \end{subfigure}
    \begin{subfigure}{0.6\textwidth}
        \includegraphics[width=\textwidth]{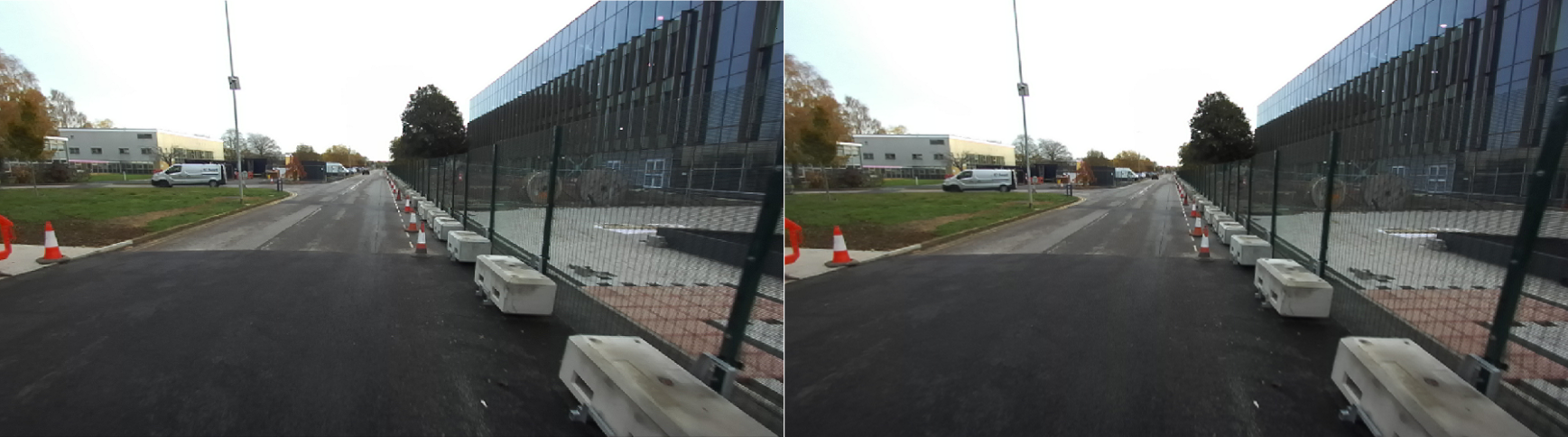}
        \caption{}
    \end{subfigure}
    \begin{subfigure}{0.3\textwidth}
        \includegraphics[width=\textwidth]{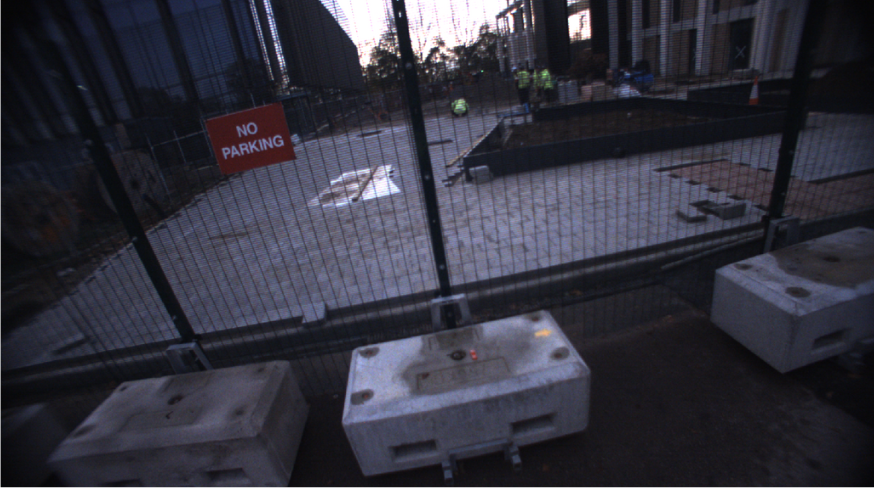}
        \caption{}
    \end{subfigure}
    
    \caption{Sensor samples from the second iteration of the backpack in \Cref{fig:iterations}b. Raw 3D scans of the side lidars are visualised in (a), (b), and combined in (c); undistorted image data from the front and rear stereo cameras are shown in (d) and (f); raw images from the monocular side cameras are seen in (e) and (g). The side cameras are synchronised with the lidars through the Pulse Per Second (PPS) signal from the GPS and a custom signal repeater; GPS data can be seen in \Cref{fig:hd_map}.}
    \label{fig:sensors}
    \vspace{-10pt}
\end{figure*}

In a second iteration, shown in \Cref{fig:iterations}b, we optimised the platform for a rugged and robust design while allowing basic height and tilt adjustments adaptable to each rider. We also focused on developing a computing unit to suit our data bandwidth requirements and finalised the wiring and cooling setup. We then run extensive system tests in the field, recordings of which can be seen in \Cref{fig:sensors}. 

Following these tests, we finalised the hardware design and the sensors' configuration and positioning as seen in \Cref{tab:Sensors} and \Cref{fig:iterations}c. By integrating noise shielding, reinforcing mounting points, and optimising weight distribution, we made the overall system more robust, compact, and comfortable for prolonged bike rides.



\section{Data Collection}
\label{sec:data}

Our goal \ref{objective.dataset} is to record a comprehensive and detailed dataset encompassing all traffic interactions and infrastructure details across diverse environmental and traffic conditions.
To achieve this, we have identified and prioritised areas of interest while remaining compliant with existing data and traffic regulations.
In particular, our data-collection strategy was formulated based on the following requirements:
\begin{description}[itemindent=0pt, leftmargin=7pt]
    \item[{\crtcrossreflabel{\textcolor{blue}{R2.1}}[req.data.dppr]}.] Ensure compliance with Data Protection and Privacy Regulations;
    \item[{\crtcrossreflabel{\textcolor{blue}{R2.2}}[req.data.road_safety]}.] Ensure compliance with Road Safety Regulations;
    \item[{\crtcrossreflabel{\textcolor{blue}{R2.3}}[req.data.diversity]}.] Capture diverse types of road layouts and different traffic volumes, road signs, cycling styles as well as various illumination, weather, and road-surface conditions;
    \item[{\crtcrossreflabel{\textcolor{blue}{R2.4}}[req.data.battery]}.] The trial length is depended on the risk assessment and must be agreed upon with the cyclists;
    \item[{\crtcrossreflabel{\textcolor{blue}{R2.5}}[req.data.bidirection]}.] Collect data bidirectionally in high- and low-peak traffic hours in specific semantically rich and diverse loops.
\end{description}

\begin{figure}[H]
    \centering

    %
    \begin{subfigure}{\columnwidth}
        \includegraphics[width=\columnwidth]{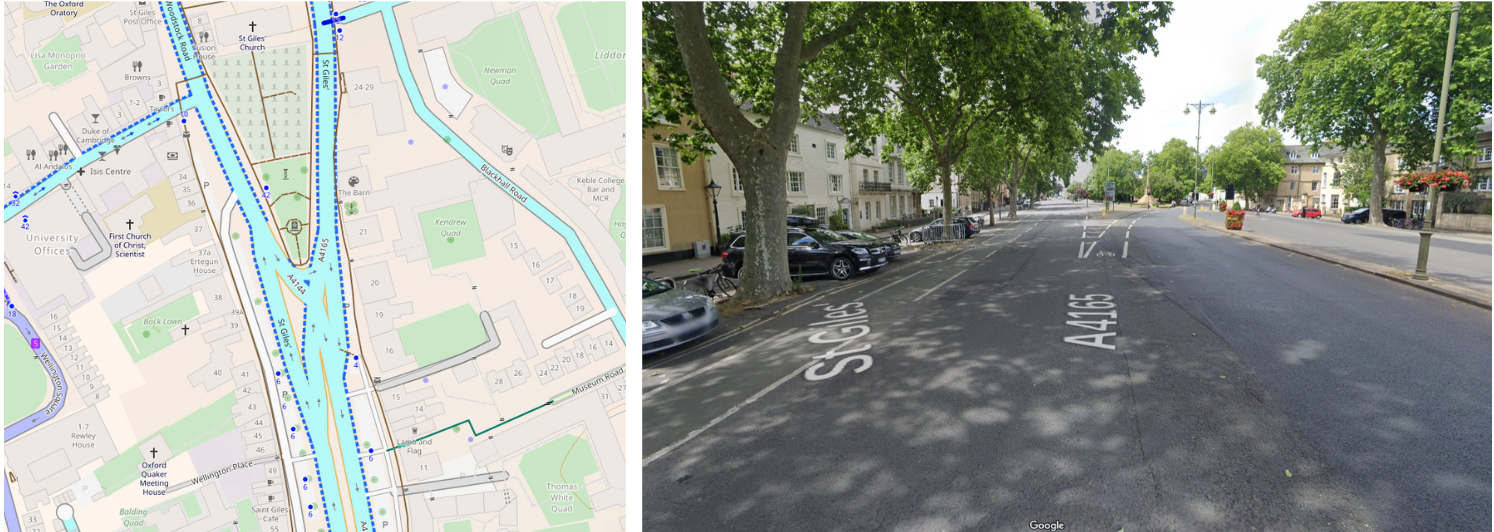}
        \caption{St. Giles' fork: Cyclists must navigate the car lanes, moving from the main bicycle lane to the lane on the right of the fork. Meanwhile, vehicles can travel and merge from either side.}
    \end{subfigure}
    \begin{subfigure}{\columnwidth}
        \includegraphics[width=\columnwidth]{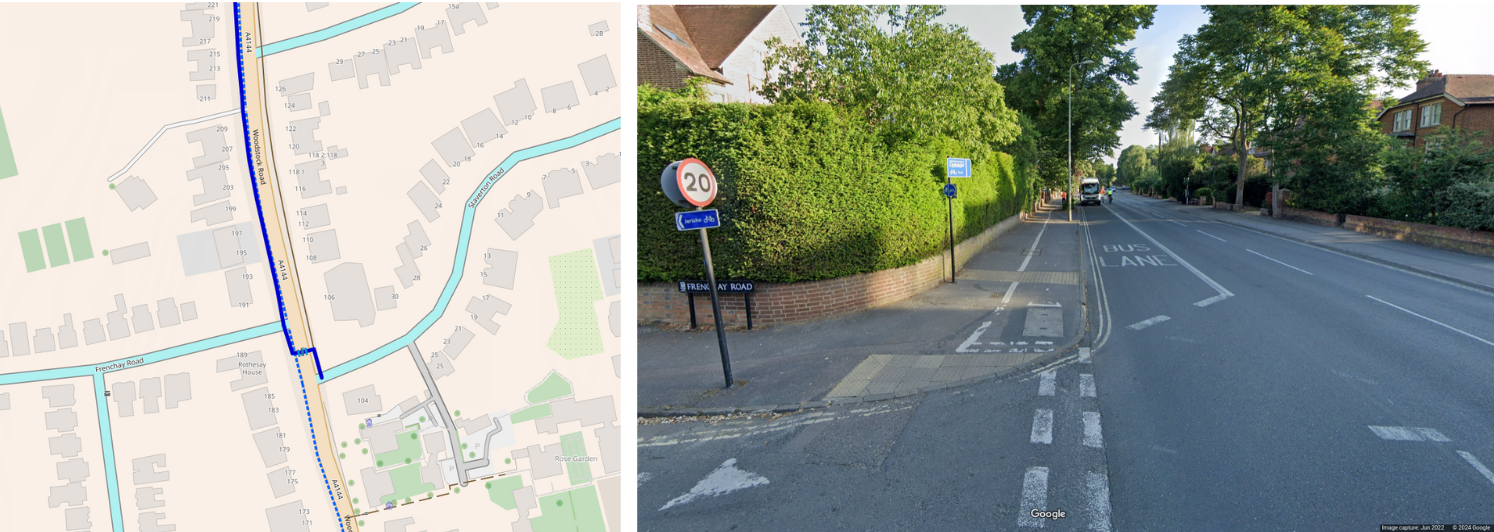}
        \caption{Woodstock Rd: The bicycle lane begins on one side of the road and later transitions to the opposite side requiring cyclists to cross the road and share the lane with pedestrians.}
    \end{subfigure}
    
    \caption[]{(Left) CyclOSM\footnotemark bicycle oriented map rendered from OpenStreetMap (OSM) \cite{OpenStreetMap}. (Right) Visualisation of places of interest on Google Maps.}
    \label{fig:routes}
    \vspace{-5pt}
\end{figure}

Oxford is known for its diverse infrastructure and road network, which has led to the collection of several extensive \gls{av} datasets \cite{maddern20171,barnes2020oxford, gadd2020sense}.
Yet, to ensure coverage of varying traffic volumes and various types of cycling infrastructure -- including roads without cycle paths, roads with shared cycle paths, and roads with exclusive cycle paths -- we conducted an extensive analysis of the city's road network.
Indeed, we first qualitatively classified Oxford's road network according to traffic, location, and functionality, also leveraging data from detailed Cycling Network Maps\footnote{\url{http://www.transportparadise.co.uk/cyclemap/}}, and then further prioritised different road segments according to the diversity of road layout, infrastructure features, and prevailing traffic conditions.
Results of this analysis can be found in \Cref{fig:routes}, leading to an estimated 15 bidirectional runs on each route, encompassing different times of the day and weather conditions (\ref{req.backpack.weight}).
The specifics of executing a single trial are contingent upon agreement with each cyclist, considering the weight of the total sensor package (\ref{req.data.bidirection}) and the cyclist's durability (\ref{req.data.battery}).

\footnotetext{\url{https://github.com/cyclosm/cyclosm-cartocss-style/}}

\section{Data Analysis}
\label{sec:analysis}
To the best of our knowledge, this is the first project capturing data from the cyclists' perspective using a multi-sensing backpack.
Our sensing platform described in \Cref{sec:backpack} provides a $360\degree$ FOV, offering a comprehensive perspective of the cyclist's surroundings, capturing dynamic and static elements in roads of diverse traffic volumes and infrastructure designed in \Cref{sec:data}.
This section describes the analysis we will perform on these data to assess, predict, and monitor the cyclists' safety on the road, as indicated in our objective \ref{objective.benchmark}.
To this end, we are focusing on building \glspl{hdmap}, data annotation, and traffic models, which will be used in conjunction to calculate a safety score that will holistically comprehend risks associated with traffic events, allowing us to address safety concerns effectively and develop strategies to mitigate them.
In this analysis, \glspl{hdmap} are central as they are a fine-grained representation of the environment, which we can augment with semantic and road quality information (smoothness, potholes, etc.) -- retrieved from \gls{imu} data -- and dynamic agents locations. Semantic elements, such as objects and local geometry, have been previously used to explain the performance and behaviour of \gls{av} systems \cite{semgat, explainable}. We aim to follow a similar approach using semantics to enhance \glspl{hdmap} and explain the behaviour of our ego-cyclist in correlation with elements in the environment.
By correlating those with traffic and trajectory models we can assess the safety of the road network.
Preliminary results are shown here on a test dataset we recorded with the proposed sensing platform at the Culham Science Center in Abingdon (UK) -- here \texttt{culham} -- and the uniD dataset \cite{unid2023}.


\subsection{HD Maps}
\label{subsec:hdmaps}
\begin{figure}[h]
    \centering
    \begin{subfigure}{0.45\textwidth}
        \includegraphics[width=\textwidth]{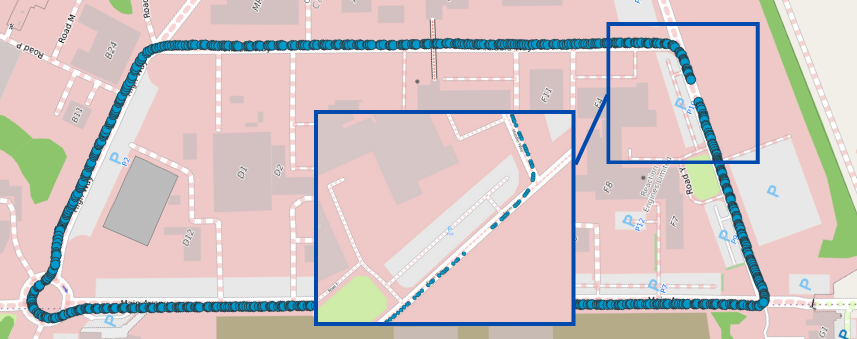}
        \caption{GPS trace overlaid onto OSM \cite{OpenStreetMap} using the on-board INS unit from \Cref{tab:Sensors}. A noticeable gap in the traces highlights moments when the GPS signal was briefly lost, providing insights into the signal interference challenges faced during data collection.}
            \vspace{10pt}
    \end{subfigure}
    \begin{subfigure}{0.45\textwidth}
        \includegraphics[width=\textwidth]{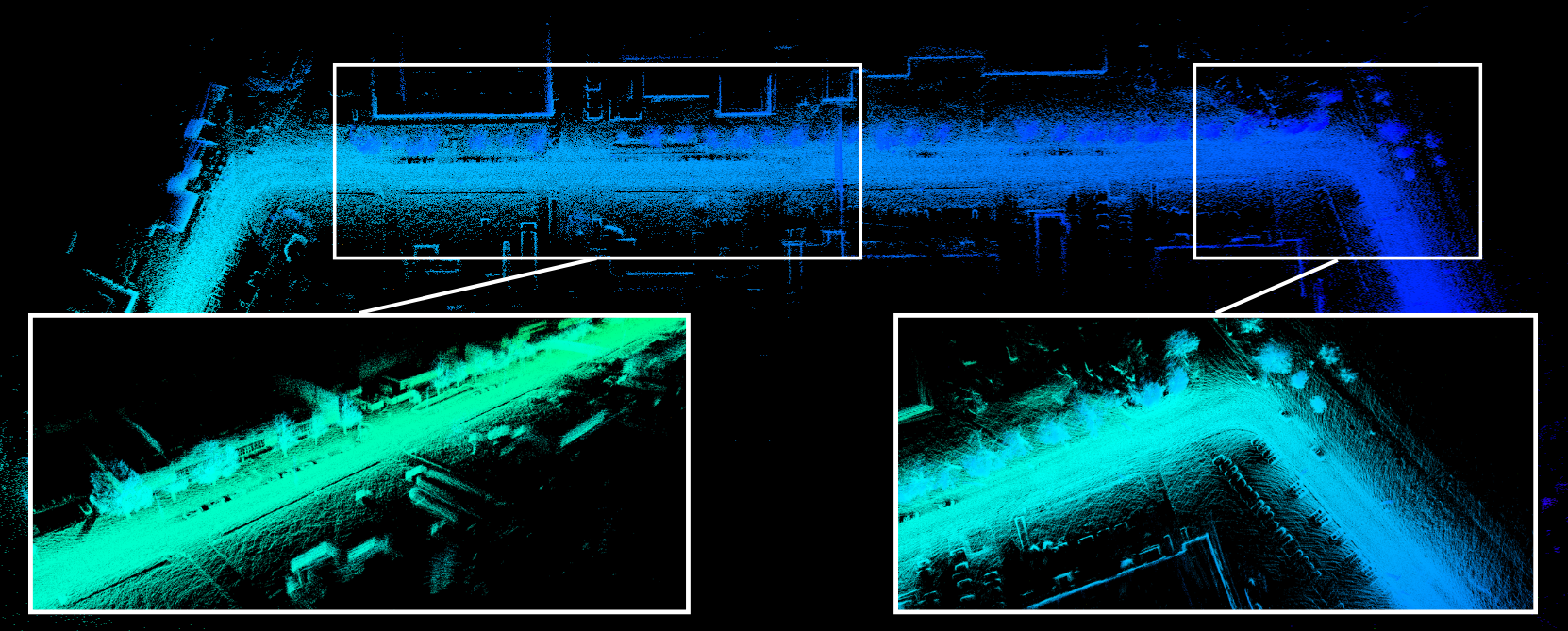}
        \caption{Preliminary results of the \gls{hdmap} generated after fusing the scans from the two side lidars sensors, generating a rich and detailed 3D structured representation of the environment.}
    \end{subfigure}
    
    \caption{GPS trace and \gls{hdmap} visualisation from our \texttt{culham} dataset.}
    \label{fig:hd_map}
\end{figure}

\Glspl{hdmap} are invaluable for providing detailed contextual information about the environment, including precise lane locations, road boundaries, curbs, and landmarks.
In addition, they capture local and global morphology, such as surface and geometry, detailing static infrastructure.
This level of accuracy offers insights into complex intersections and infrastructure, aiding in localising all road agents and allowing us to correlate their behaviours to traffic flows and interactions.
A preliminary example of an \gls{hdmap} applied to \texttt{culham} is depicted in \Cref{fig:hd_map}. 

\subsection{Data Annotation}
\label{subsec:annotation}
\begin{figure}[]
    \centering
    \begin{subfigure}{0.44\textwidth}
        \includegraphics[width=\textwidth]{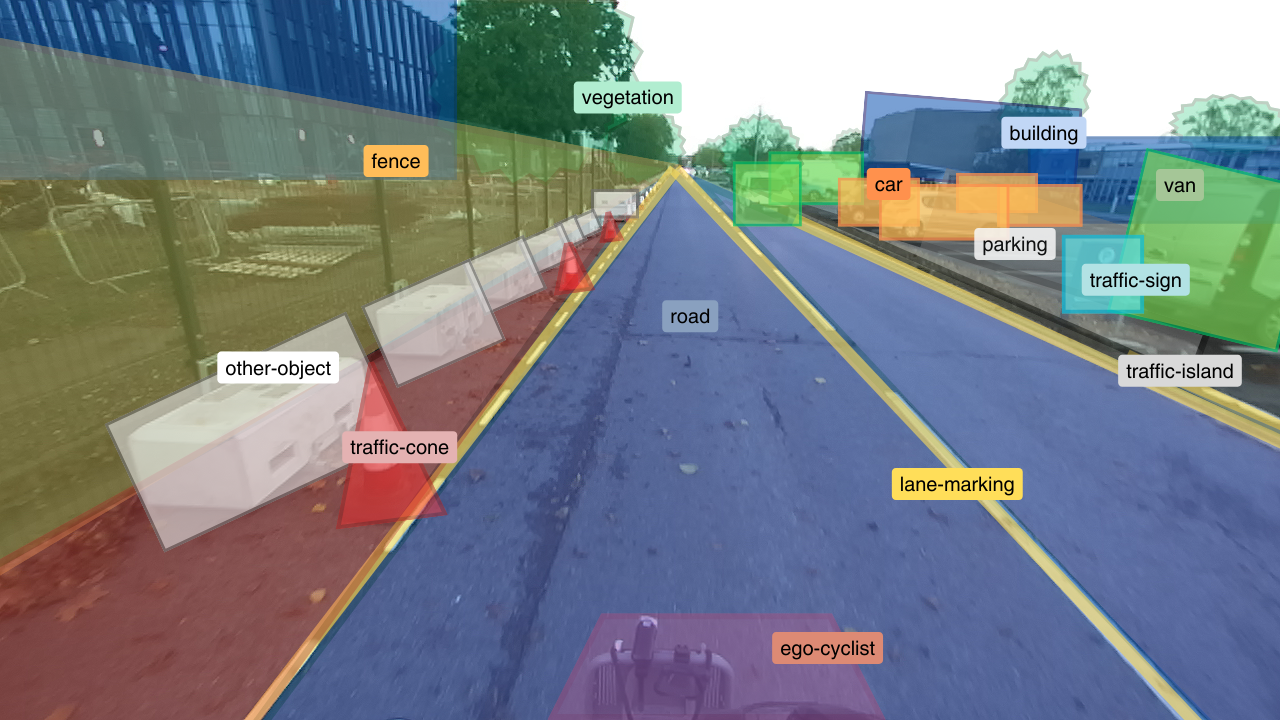}
        \caption{}
    \end{subfigure}
    \begin{subfigure}{0.44\textwidth}
        \includegraphics[width=\textwidth]{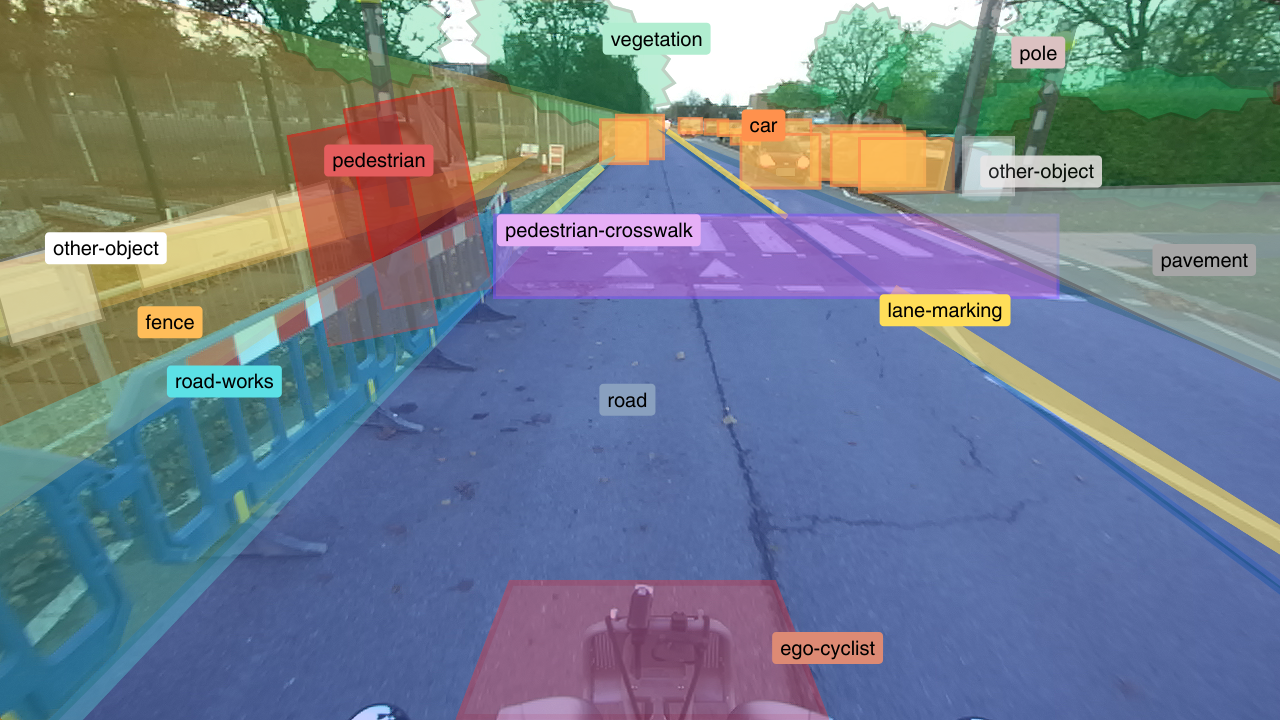}
        \caption{}
    \end{subfigure}
    
    \caption{Sample labelled images from two locations depicting different traffic scenarios in our \texttt{culham} dataset. Label categories can be seen in \Cref{tab:labels}.}
    \label{fig:labels}
    \vspace{-15pt}
\end{figure}

To analyse the data and extract trajectory and behavioural patterns, we need to detect and track semantics of interest in the dataset.
We will annotate all dynamic agents and road-layout-specific classes in the dataset, extending existing class definitions -- e.g. the extensively-used CityScapes' \cite{cordts2016cityscapes} -- including infrastructure elements critical to a thorough description of the scene.
\Cref{tab:labels} summarises our semantic class definition of interest and \Cref{fig:labels} depicts labelled images in our \texttt{culham} dataset.

\begin{table}[]
    \centering
    \renewcommand{\arraystretch}{1.3}

\begin{tabular}{|l|p{0.65\columnwidth}|}
    \hline
    \textbf{Category} & \textbf{Subcategories} \\
    \hline
    \textbf{ground} & road, pavement, cycle-lane, bus-lane, parking-spot, bus-stop, lane-marking, roundabout, crossing, traffic-island, other \\
    \hline
    \textbf{structure} & building, fence, wall, barrier, stop-shelter, road-works, overhead-bridge, other \\
    \hline
    \textbf{nature} & vegetation, trunk \\
    \hline
    \textbf{object} & pushchair, sign, fence, pole, traffic-cone, traffic-light, wheelchair, other \\
    \hline
    \textbf{vehicle} & car, truck, bicycle, motorcycle, e-scooter, mobility-scooter, van, bus, on-rails, other \\
    \hline
    \textbf{animate-agent} & animal, pedestrian, rider, other\\
    \hline
\end{tabular}
\caption{Target categories and derived subcategories for semantic labelling of image and pointcloud data focusing on infrastructure elements and dynamic agents.}
    \label{tab:labels}
\vspace{-0.5cm}
\end{table}

\subsection{Traffic and Trajectory Models}
\label{subsec:trafficmodels}
Since this dataset will be taken from the perspective of a \gls{vru} instead of a vehicle, it will contain diverse VRU-to-VRU and VRU-to-Vehicle interactions.
These interactions have higher variability as \glspl{vru} do not necessarily follow allocated lanes, traffic lights, and road rules.
Especially cyclists, because they operate at higher speeds, pose a danger to other types of road users, making predicting cyclist trajectories crucial for a comprehensive safety assessment.

\begin{figure*}[h]
    \centering
    \begin{subfigure}{\columnwidth}
        \includegraphics[width=\textwidth]{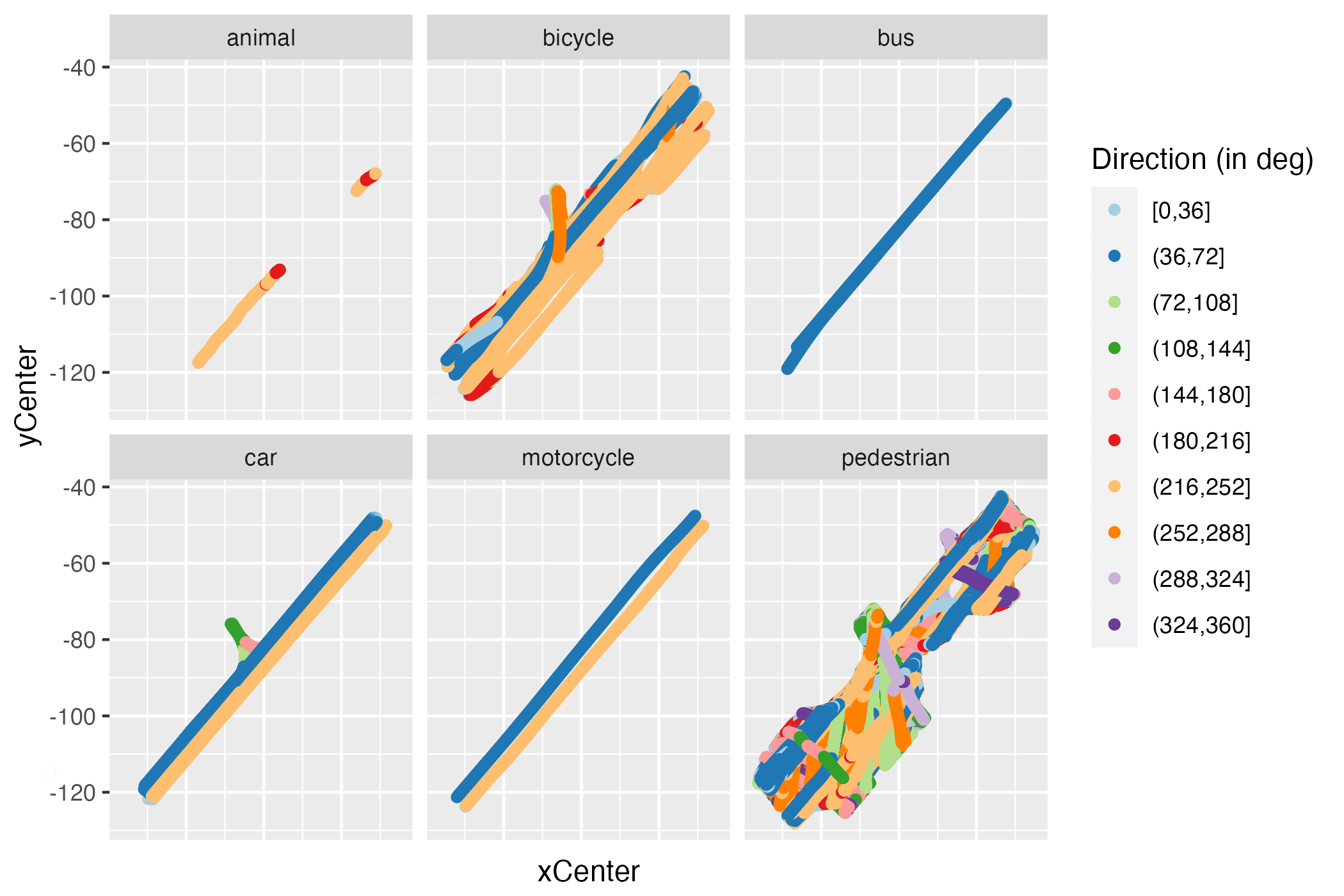}
        \label{fig:traffic:a}
    \end{subfigure}
    \hfill
    \begin{subfigure}{\columnwidth}
        \includegraphics[width=\textwidth]{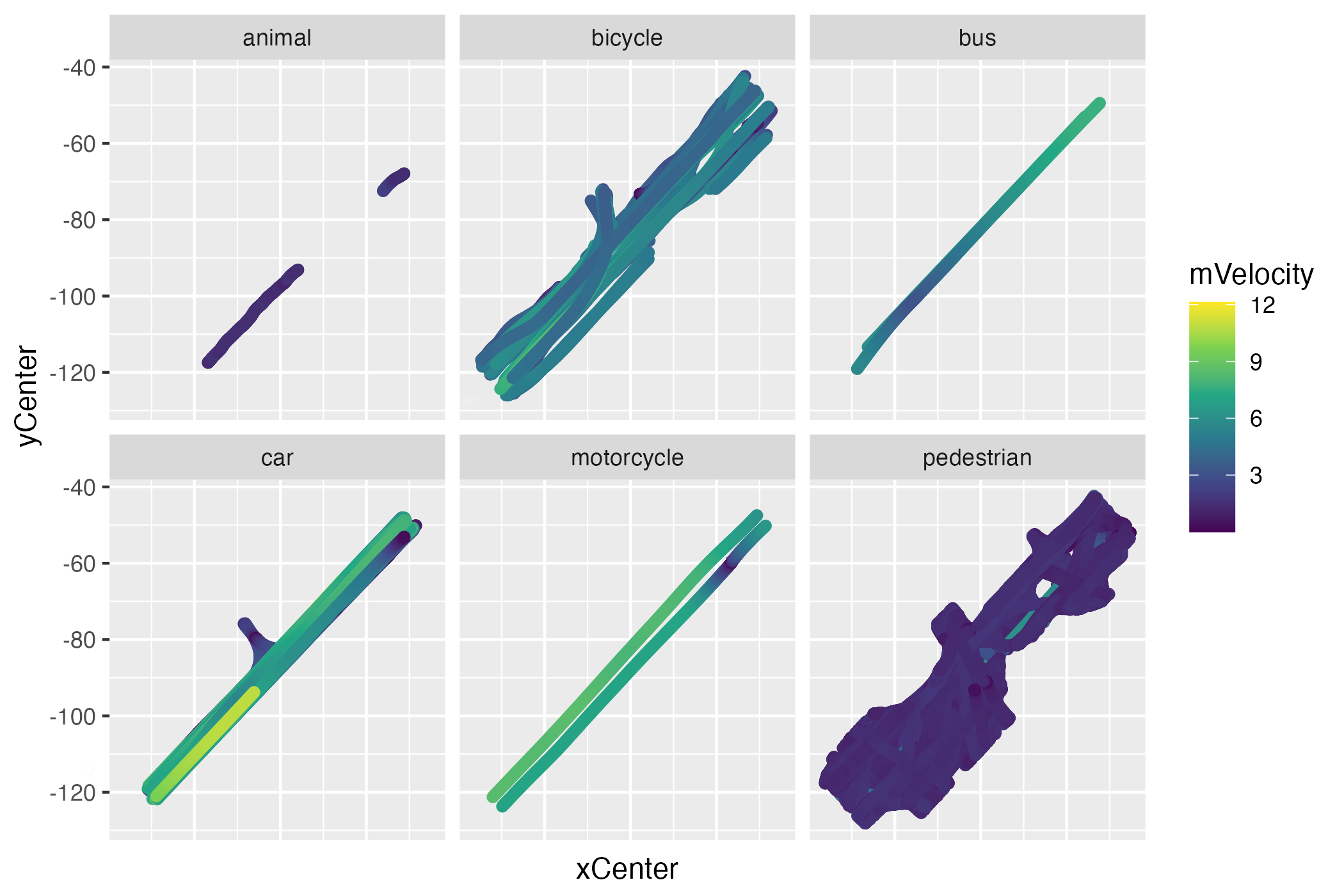}
        \label{fig:traffic:b}
    \end{subfigure}
\vspace{-10pt}
    \caption{Preliminary analysis results for road agents' direction (a) and velocity (b) on the uniD dataset \cite{unid2023}. Interestingly, \glspl{vru} exhibit a  more varied behaviour w.r.t. engine-powered vehicles, which follow road lanes more closely.}
    \label{fig:trafficmodelling}
    
\end{figure*}

We have begun developing these processes by leveraging the uniD dataset \cite{unid2023}.
\Cref{fig:trafficmodelling} shows results for road agents' tracked directions and velocity, divided by kind.
As expected, \glspl{vru} tend to occupy a larger portion of the road surface exhibiting a wider range of behaviours.

%

\section{Future Work and Open Issues}
\label{sec:discussion}
As the outlined work packages have provided encouraging preliminary results, further refinement and analysis are imperative for the project's success.
We aim to conduct a data-driven safety analysis leveraging behavioural modelling and trajectory prediction methods from AV research.
In addition, an in-depth causality analysis is required to identify the intricate causes and effects of each event captured.

Our initial data collection is limited to private and constrained environments, capturing very few driving interactions; in the final iteration of our project, we will collect the Oxford dataset described in this paper, which we commit to making publicly available.
Using this novel dataset, we aim to develop a generalisable approach to assessing safety, which, based on infrastructure, traffic volumes, and historical data, can generate insights into potentially unsafe situations, even on roads not traversed within the dataset. Within the scope of this project, our focus lies on urban environments. However, an extension to off-road areas can provide valuable insights into the behaviour of \glspl{vru} across more diverse and challenging environments  \cite{gadd2024oord}.

\section{Conclusion}
\label{sec:conclusion}
This paper introduces the RobotCycle project which aims to evaluate the safety of cycling infrastructure by leveraging advancements in AV research and technologies. Here, we have described the development process of a wearable data collection device, the design of the data collection process, and the data analysis rationale.
Our primary focus has been on building and testing the sensing backpack to ensure that it satisfies all design requirements so that we can achieve the project's objectives.
Initial results demonstrate the device's ability to capture diverse and rich datasets, which are then used to extract semantic elements, \glspl{hdmap}, and traffic patterns.
The RobotCycle project is an innovative undertaking with a mission to enhance the safety and sustainability of urban cities.
We aim to generate valuable insights into the ongoing discourse on urban development, setting a benchmark for further advancements in the field. 

\bibliographystyle{IEEEtran}
\bibliography{biblio}

\begin{thebibliography}{10}
\providecommand{\url}[1]{#1}
\csname url@samestyle\endcsname
\providecommand{\newblock}{\relax}
\providecommand{\bibinfo}[2]{#2}
\providecommand{\BIBentrySTDinterwordspacing}{\spaceskip=0pt\relax}
\providecommand{\BIBentryALTinterwordstretchfactor}{4}
\providecommand{\BIBentryALTinterwordspacing}{\spaceskip=\fontdimen2\font plus
\BIBentryALTinterwordstretchfactor\fontdimen3\font minus \fontdimen4\font\relax}
\providecommand{\BIBforeignlanguage}[2]{{%
\expandafter\ifx\csname l@#1\endcsname\relax
\typeout{** WARNING: IEEEtran.bst: No hyphenation pattern has been}%
\typeout{** loaded for the language `#1'. Using the pattern for}%
\typeout{** the default language instead.}%
\else
\language=\csname l@#1\endcsname
\fi
#2}}
\providecommand{\BIBdecl}{\relax}
\BIBdecl

\bibitem{Shoman2023}
M.~M. Shoman, H.~Imine, E.~M. Acerra, and C.~Lantieri, ``{Evaluation of Cycling Safety and Comfort in Bad Weather and Surface Conditions Using an Instrumented Bicycle},'' \emph{IEEE Access}, vol.~11, no. January, pp. 15\,096--15\,108, 2023.

\bibitem{Kaparias2021}
I.~Kaparias, S.~Miah, S.~Clegg, Y.~Gao, B.~Waterson, and E.~Milonidis, ``{Measuring the effect of highway design features on cyclist behavior using an instrumented bicycle},'' \emph{2021 7th International Conference on Models and Technologies for Intelligent Transportation Systems, MT-ITS 2021}, pp. 1--6, 2021.

\bibitem{Vieira2016}
P.~Vieira, J.~P. Costeira, S.~Brandao, and M.~Marques, ``{SMARTcycling: Assessing cyclists' driving experience},'' \emph{IEEE Intelligent Vehicles Symposium, Proceedings}, vol. 2016-August, no.~Iv, pp. 1321--1326, 2016.

\bibitem{Ibrahim2021a}
M.~R. Ibrahim, J.~Haworth, N.~Christie, and T.~Cheng, ``{CyclingNet: Detecting cycling near misses from video streams in complex urban scenes with deep learning},'' \emph{IET Intelligent Transport Systems}, vol.~15, no.~10, pp. 1331--1344, 2021.

\bibitem{Kumar2023}
S.~M. Kumar, ``{Smart Biking : IoT-Connected Cycling Gear for Training and Safety},'' \emph{2023 Second International Conference On Smart Technologies For Smart Nation (SmartTechCon)}, pp. 652--656, 2023.

\bibitem{Wen2016}
C.~Wen, S.~Pan, C.~Wang, and J.~Li, ``{An Indoor Backpack System for 2-D and 3-D Mapping of Building Interiors},'' \emph{IEEE Geoscience and Remote Sensing Letters}, vol.~13, no.~7, pp. 992--996, 2016.

\bibitem{Gong2021}
Z.~Gong, J.~Li, Z.~Luo, C.~Wen, C.~Wang, and J.~Zelek, ``{Mapping and Semantic Modeling of Underground Parking Lots Using a Backpack LiDAR System},'' \emph{IEEE Transactions on Intelligent Transportation Systems}, vol.~22, no.~2, pp. 734--746, 2021.

\bibitem{Chen2021}
P.~Chen, W.~Shi, S.~Bao, M.~Wang, W.~Fan, and H.~Xiang, ``{Low-Drift Odometry, Mapping and Ground Segmentation Using a Backpack LiDAR System},'' \emph{IEEE Robotics and Automation Letters}, vol.~6, no.~4, pp. 7285--7292, 2021.

\bibitem{Nuchter2015}
A.~N{\"{u}}chter, D.~Borrmann, P.~Koch, M.~K{\"{u}}hn, and S.~May, ``{A Man-Portable, Imu-Free Mobile Mapping System},'' \emph{ISPRS Annals of the Photogrammetry, Remote Sensing and Spatial Information Sciences}, vol.~2, no. 3W5, pp. 17--23, 2015.

\bibitem{Laguela2018}
S.~Lag{\"{u}}ela, I.~Dorado, M.~Gesto, P.~Arias, D.~Gonz{\'{a}}lez-Aguilera, and H.~Lorenzo, ``{Behavior analysis of novel wearable indoor mapping system based on 3d-slam},'' \emph{Sensors (Switzerland)}, vol.~18, no.~3, pp. 1--16, 2018.

\bibitem{Corso2013}
N.~Corso and A.~Zakhor, ``{Indoor localization algorithms for an ambulatory human operated 3D mobile mapping system},'' \emph{Remote Sensing}, vol.~5, no.~12, pp. 6611--6646, 2013.

\bibitem{Holmgren2017}
J.~Holmgren, H.~M. Tulldahl, J.~Nordl{\"{o}}f, M.~Nystr{\"{o}}m, K.~Olofsson, J.~Rydell, and E.~Willen, ``{Estimation of tree position and stem diameter using simultaneous localization and mapping with data from a backpack-mounted laser scanner},'' \emph{International Archives of the Photogrammetry, Remote Sensing and Spatial Information Sciences - ISPRS Archives}, vol.~42, no. 3W3, pp. 59--63, 2017.

\bibitem{Rasch2022}
A.~Rasch and M.~Dozza, ``{Modeling Drivers' Strategy When Overtaking Cyclists in the Presence of Oncoming Traffic},'' \emph{IEEE Transactions on Intelligent Transportation Systems}, vol.~23, no.~3, pp. 2180--2189, 2022.

\bibitem{Daraei2021}
S.~Daraei, K.~Pelechrinis, and D.~Quercia, ``{A data-driven approach for assessing biking safety in cities},'' \emph{EPJ Data Science}, vol.~10, no.~1, 2021.

\bibitem{Castells-Graells2020}
D.~Castells-Graells, C.~Salahub, and E.~Pournaras, ``{On cycling risk and discomfort: urban safety mapping and bike route recommendations},'' \emph{Computing}, vol. 102, no.~5, pp. 1259--1274, 2020.

\bibitem{Meuleners2023}
L.~Meuleners, M.~Fraser, and P.~Roberts, ``{Improving cycling safety through infrastructure design: A bicycle simulator study},'' \emph{Transportation Research Interdisciplinary Perspectives}, vol.~18, no. June 2022, p. 100768, 2023.

\bibitem{Ibrahim2021b}
\BIBentryALTinterwordspacing
M.~R. Ibrahim, J.~Haworth, and N.~Christie, ``{Re-designing cities with conditional adversarial networks},'' 2021. [Online]. Available: \url{http://arxiv.org/abs/2104.04013}
\BIBentrySTDinterwordspacing

\bibitem{maddern20171}
W.~Maddern, G.~Pascoe, C.~Linegar, and P.~Newman, ``1 year, 1000 km: The oxford robotcar dataset,'' \emph{The International Journal of Robotics Research}, vol.~36, no.~1, pp. 3--15, 2017.

\bibitem{barnes2020oxford}
D.~Barnes, M.~Gadd, P.~Murcutt, P.~Newman, and I.~Posner, ``The oxford radar robotcar dataset: A radar extension to the oxford robotcar dataset,'' in \emph{2020 IEEE International Conference on Robotics and Automation (ICRA)}.\hskip 1em plus 0.5em minus 0.4em\relax IEEE, 2020, pp. 6433--6438.

\bibitem{gadd2020sense}
M.~Gadd, D.~De~Martini, L.~Marchegiani, P.~Newman, and L.~Kunze, ``Sense--assess--explain (sax): Building trust in autonomous vehicles in challenging real-world driving scenarios,'' in \emph{2020 IEEE Intelligent Vehicles Symposium (IV)}.\hskip 1em plus 0.5em minus 0.4em\relax IEEE, 2020, pp. 150--155.

\bibitem{OpenStreetMap}
{OpenStreetMap contributors}, ``{Planet dump retrieved from https://planet.osm.org },'' \url{ https://www.openstreetmap.org }, 2017.

\bibitem{semgat}
E.~Panagiotaki, D.~De~Martini, G.~Pramatarov, M.~Gadd, and L.~Kunze, ``Sem-gat: Explainable semantic pose estimation using learned graph attention,'' in \emph{2023 21st International Conference on Advanced Robotics (ICAR)}, 2023, pp. 367--374.

\bibitem{explainable}
E.~Panagiotaki, D.~De~Martini, and L.~Kunze, ``Semantic interpretation and validation of graph attention-based explanations for gnn models,'' in \emph{2023 21st International Conference on Advanced Robotics (ICAR)}, 2023, pp. 375--380.

\bibitem{unid2023}
\BIBentryALTinterwordspacing
LevelXData, ``The unid dataset: Naturalistic trajectories of vehicles and vulnerable road users recorded at the rwth aachen university campus,'' Tech. Rep., 2023. [Online]. Available: \url{https://levelxdata.com/unid-dataset/}
\BIBentrySTDinterwordspacing

\bibitem{cordts2016cityscapes}
M.~Cordts, M.~Omran, S.~Ramos, T.~Rehfeld, M.~Enzweiler, R.~Benenson, U.~Franke, S.~Roth, and B.~Schiele, ``The cityscapes dataset for semantic urban scene understanding,'' in \emph{Proceedings of the IEEE conference on computer vision and pattern recognition}, 2016, pp. 3213--3223.

\bibitem{gadd2024oord}
M.~Gadd, D.~De~Martini, O.~Bartlett, P.~Murcutt, M.~Towlson, M.~Widojo, V.~Mu{\c{s}}at, L.~Robinson, E.~Panagiotaki, G.~Pramatarov \emph{et~al.}, ``Oord: The oxford offroad radar dataset,'' \emph{arXiv preprint arXiv:2403.02845}, 2024.

\end{thebibliography}

\end{document}